\documentclass[runningheads, orivec]{llncs}
\providecommand{\authcount}[1]{\vspace{-0.5cm}}
\usepackage[a4paper, margin=3cm]{geometry}
\usepackage{titletoc}
\usepackage{hyperref}
\usepackage[page,header]{appendix}

\usepackage[T1]{fontenc}
\usepackage{graphicx}
\usepackage{booktabs}
\usepackage[misc]{ifsym}
\newcommand{\corr}{(\Letter)}

\usepackage{algorithm} 
\usepackage{comment} 
\usepackage{algpseudocode}
\usepackage{tikz} 

\usepackage{amsmath}
\usepackage{booktabs}       
\usepackage{amsfonts}       
\usepackage{nicefrac}       
\usepackage{mathabx}
\usepackage{cite}
\usepackage{subcaption}

\definecolor{darkpink}{RGB}{200, 0, 100}

\newcommand{\KL}{D_{\mathrm{KL}}}
\newcommand{\R}{\mathbb{R}}
\newcommand{\dd}{\mathop{}\mathopen{}\mathrm{d}}

\usepackage{microtype}      

\makeatletter
\DeclareRobustCommand{\iscircle}{\mathord{\mathpalette\is@circle\relax}}
\newcommand\is@circle[2]{%
  \begingroup
  \sbox\z@{\raisebox{\depth}{$\m@th#1\bigcirc$}}%
  \sbox\tw@{$#1\square$}%
  \resizebox{!}{\ht\tw@}{\usebox{\z@}}%
  \endgroup
}

\makeatother

\begin{document}

\title{\texorpdfstring{Exposing the Illusion of Fairness: \\ Auditing Vulnerabilities to Distributional Manipulation Attacks}{Exposing the Illusion of Fairness: - Auditing Vulnerabilities to Distributional Manipulation Attacks}}

\titlerunning{Exposing the Illusion of Fairness}

\author{
Valentin Lafargue\inst{1,2,3,4} \corr 
\and Adriana Laurindo Monteiro \inst{5} 
\and Emmanuelle Claeys \inst{4, 6} 
\and Laurent Risser \inst{1, 3}
\and Jean-Michel Loubes \inst{2,3}
}

\institute{
Institut de Mathématiques de Toulouse (IMT), Toulouse, France 
\and
Institut national de recherche en sciences et technologies du numérique (INRIA), Bordeaux, France \email{valentin.lafargue@math.univ-toulouse.fr}
\and
Artificial and Natural Intelligence Toulouse Institute 2 (ANITI 2), Toulouse, France
\and
Institut de Recherche en Informatique de Toulouse (IRIT), Toulouse, France
\and
Independent Researcher
\and 
CNRS IRL CROSSING, Adelaide, Australia
}

\authorrunning{Lafargue et al.}

\maketitle
\begin{abstract}
The rapid deployment of AI systems in high-stakes domains, including those classified as high-risk under the The EU AI Act (Regulation (EU) 2024/1689), has intensified the need for reliable compliance auditing. For binary classifiers, regulatory risk assessment often relies on global fairness metrics such as the Disparate Impact ratio, widely used to evaluate potential discrimination. In typical auditing settings, the auditee provides a subset of its dataset to an auditor, while a supervisory authority may verify whether this subset is representative of the full underlying distribution. In this work, we investigate to what extent a malicious auditee can construct a fairness-compliant yet representative-looking sample from a non-compliant original distribution, thereby creating an illusion of fairness. We formalize this problem as a constrained distributional projection task and introduce mathematically grounded manipulation strategies based on entropic and optimal transport projections. These constructions characterize the minimal distributional shift required to satisfy fairness constraints. To counter such attacks, we formalize representativeness through distributional distance based statistical tests and systematically evaluate their ability to detect manipulated samples. Our analysis highlights the conditions under which fairness manipulation can remain statistically undetected and provides practical guidelines for strengthening supervisory verification. We validate our theoretical findings through experiments on standard tabular datasets for bias detection. Code is publicly available at 
\url{https://github.com/ValentinLafargue/Inspection}.


\end{abstract}

\section{Introduction}

Fairness auditing has emerged as a critical practice to ensure that machine learning models comply with ethical and legal standards by not exhibiting discriminatory bias \cite{barocas-hardt-narayanan,disparate,oneto2020fairness,wang2022brief}. High-profile investigative audits, such as the ProPublica analysis of the COMPAS recidivism risk tool, have exposed significant biases against certain demographic groups~\cite{angwin2016machine}. These findings underscored the societal harms of unverified AI systems and prompted calls for regular fairness audits by independent parties~\cite{raji2020closing}. 

In response, regulators have begun instituting fairness compliance requirements. The Regulation (EU) 2024/1689 establishes strict requirements on high-risk AI systems. Under Article 9, providers must implement a comprehensive risk management system that identifies, analyzes, and evaluates risks related to the system’s intended use and ensure that any residual risk is reduced to an acceptable level before deployment. 

Auditing an algorithm is a long multidimensional process that aims to verify a long list of properties in order to establish regulatory compliance. In this paper, we focus on a specific part of this long pipeline process: evaluating whether the AI system may induce discriminatory outcomes with respect to regulator-designated sensitive attributes. We refer to this objective as fairness compliance. In the United States, such assessments are commonly operationalized through disparate impact analysis, which evaluates indirect discrimination in algorithms to a legally protected group~\cite{feldman2015certifying}. One practical strategy to operationalize such obligations is outcome-based evaluation. In the US, this is commonly implemented through disparate impact analysis in EEOC guidelines \cite{Uni_guidelines}, and supported by case law (e.g., Griggs v. Duke Power Co., 401 U.S. 424 (1971)). In this case, the US Supreme Court ruled illegal a hiring process if it results in disparate impact; even if the hiring decision is not calculated based on the protected attribute.

The auditee chooses a sample of the dataset on which the fairness measure is estimated. Hence, ensuring the good faith of the auditee is critical, as exemplified by the Volkswagen emissions scandal \cite{Volkswagen}, in which vehicles were engineered to detect auditing conditions and deceive regulators by temporarily producing compliant behavior. Similar concerns arise in machine learning, where inconsistencies in model behavior during auditing have been documented, such as in the case of Meta's models discussed in \cite{bourree2025robustmlauditingusing}. 

In this work, our objective is to support auditors by identifying potential strategies that audited entities might use to circumvent fairness audits and by providing tools to detect such attempts. Building on the notion of \textit{manipulation-proof} introduced in \cite{pmlr-v162-yan22c}, we show how a dataset that initially violates a fairness criterion, such as Disparate Impact, can be minimally altered to appear compliant, with limited distributional shift as measured by the Kullback–Leibler (KL) divergence or the Wasserstein distance. By systematically analyzing these plausible manipulations, we aim to raise awareness of audit vulnerabilities and to equip oversight bodies with methods to detect suspicious modifications, thereby strengthening the reliability and robustness of fairness auditing processes.

\begin{figure}[t!]
  \begin{centering}
      \include{Images/Mathcha/Highest_unbiasedness_2}
  \caption{
  Highest undetected achievable Disparate Impact for 7 dataset samplings (10\%) across all fair-washing methods. Samples are considered manipulated if they fail a single statistical test for distributional equivalence (KL divergence (KL), Wasserstein distance (W),  Kolmogorov–Smirnov (KS) or Maximum Mean Discrepancy (MMD) at a 95\% confidence level).  }
  \label{fig:Highest_unbiasedness}
  \end{centering}
\end{figure}  

Our contributions are the following:
\begin{itemize}
    \item We present a mathematical analysis of data manipulation aimed at artificially improving fairness. Our framework considers distributional shifts of the evaluation sample designed to satisfy fairness constraints while remaining statistically indistinguishable from the original distribution. We study two approaches: entropic projections and optimal-transport projections.
    \item We provide a testing methodology to assess whether, and to what extent, a sample from the projected distribution can artificially increase the  Disparate Impact, thereby faking fairness, while remaining undetected by distribution-based statistical tests (Figure~\ref{fig:Highest_unbiasedness}).
\end{itemize}

\section{Problem statement}

Our auditing framework involves three entities:
\begin{enumerate}
    \item \textbf{Auditee.} The product owner holds the full dataset and trained predictor $f$ (binary classifier, $\widehat{Y}\in\{0,1\}$). For regulatory or sovereignty reasons, only a subset of the data is disclosed to the auditor. The auditee is liable (i) if the model is not fairness-compliant and (ii) if the submitted subset does not enable the auditor to evaluate the fairness of the system.
    \item \textbf{Auditor.} The auditor is an external entity in charge of computing prescribed fairness metrics and determining whether the system satisfies the relevant compliance criteria. The auditor’s evaluation is conducted solely on the basis of the information disclosed by the auditee. In particular, using the empirical distribution of the submitted sample, it estimates a fairness metric. In most of the paper we consider the example where the prescribed metric is the Disparate Impact Ratio.
    \item \textbf{Supervisory authority.} A higher-level oversight body, such as a regulator acting as a market surveillance authority, for instance the ACPR for banking regulations, or a court-appointed expert intervening in juridical cases or regulatory verification. It may intervene in regulatory reviews or judicial proceedings. In particular it has access to the full dataset, and can (i) compute fairness metric on full data and (ii) determine subset representativeness. It is responsible for ensuring that the audit was conducted correctly and in accordance with applicable requirements.
\end{enumerate}

In this paper we consider a use case where the auditee has developed a model which has fairness issues. It tries to hide the problem by picking a subsample while optimizing the fairness metric that will be computed by the auditor. Yet, from the supervisory authority, submitting a non-representative sample constitute a deceptive attempt by the auditee to obstruct or distort the assessment.

\subsection{Auditor fairness evaluation}

One of the most desired fairness properties is the statistical parity, which ensures that the decision of the algorithm does not depend on the sensitive attribute. We note $S$ the protected attribute, also called sensitive attribute in the machine learning literature. In this paper, the protected attribute will always be binary \\ $S\in\{0,1\}$, this is standard to observe treatment disparity for gender, disability or ethnicity (one-versus-all paradigm) ; continuous variable such as income or age are transformed to binary using a threshold value as separator (i.e., above or below 40 years old).  The auditor will apply a fairness metric, the Disparate Impact (DI) ratio, to evaluate the statistical parity of the subset provided by the audited entity ; defined for a model $\widehat{Y}=f(X)$ and data following distribution $\mathbb{P}$ by 

\begin{equation}
    \label{DI}
    DI(f,\mathbb{P}, S):= \frac{\min(\mathbb{P}(\widehat{Y} = 1 \mid S = 0), \mathbb{P}(\widehat{Y} = 1 \mid S = 1)}{\max(\mathbb{P}(\widehat{Y} = 1 \mid S = 0), \mathbb{P}(\widehat{Y} = 1 \mid S = 1)}.
\end{equation}

This quantity is equal to 1 when no probabilistic relationship exists between the outcome of the model and the sensitive variable, which implies a strict independence in the case where $f(X)$ is a two-class classification model. 
Hence, several norms or regulations impose that a model should have its disparate impact ratio greater than a given level $t$, often set to $t=0.8$ \cite{Uni_guidelines, feldman2015certifying, wright2024null}. 
While this work focuses on the DI ratio, our methods and results stands for other common global fairness metric, see in the Appendix, Sec.~\ref{app:sec:other_fairness_metric}.

\subsection{Supervisory verification}

Let $(E,\mathcal{B}(E))$ be a measurable space. Denote by $\mathcal{P}(E)$ and $\mathcal{M}(E)$, respectively the space of probability measures on $E$ and the space of finite measures in $E$. Let $Q_n =\frac{1}{n} \sum_{i=1}^n \delta_{Z_i}$ be the empirical underlying distribution of the auditee where $Z_i$ is an i.i.d. sample of a random variable with value in $E$, $\delta_{Z_i}$ is the Dirac measure associated ; and let $\widehat{Q}_{m}$ be the empirical distribution from the auditee sample $\mathcal{D}_m$ (with $m \leq n)$. The supervisory goal is the verification that  $\widehat{Q}_{m}$ is representative of $Q_n$. 
Consider a distance $d$ in $E$, for $\widehat{Q}_{m}$ to be representative of $Q_n$, it needs to be close to $Q_n$: $d(\widehat{Q}_{m},Q_n)$ needs to be small.

\subsection{Malicious auditee optimization problem} \label{s:method}

We propose a methodology that simulates how stakeholders could try to evade an audit on the Disparate Impact ratio, without any liability. The auditee aims to construct a distribution optimally close to the distribution of the original data, while ensuring that the fairness measure is above a threshold, as required by the regulations. In reality, the construction of a falsely compliant dataset is modeled as finding the solution to the optimization problem : 
\begin{equation}
    \mathrm{argmin}_{P\in \mathcal{P}(E), DI(f,P,S)  \geq t} d(P,Q_n).
\end{equation}

Concerning which distance $d$ to minimize, the malicious auditee will choose depending on the supervisory authority verification strategy. 

\begin{remark}
The auditee is responsible to provide a representative subset on which the auditor will compute the fairness measure. The compliance result will be determined on this subset, which  should be chosen to convey the general properties of the data of the auditee.
\end{remark}

\section{Related Work}

Fairness auditing has increasingly been shown to be vulnerable to manipulation, commonly referred to as \emph{fairwashing} \cite{aivodji2019fairwashing}. A key vulnerability stems from the fact that global fairness metrics, such as Disparate Impact or Demographic Parity, are estimated on a test sample and therefore depend on the audited data distribution. Exploiting this dependence, Fukuchi et al. introduced stealthily biased sampling, showing how curated samples can appear fair while remaining close to the original biased distribution \cite{fukuchi2020faking}. Proxy based attacks further demonstrate that interpretable surrogate models can be presented as evidence of fairness while masking discriminatory behavior \cite{aivodji2019fairwashing}. Other works have studied decision manipulation during audits \cite{lemerrer2020faking}, empirical inconsistencies between audited and deployed behavior \cite{raji2020closing}, and theoretical limits of detecting such manipulation \cite{shamsabadi2023fairwashing}.

Several fair washing strategies are closely related to classical bias mitigation techniques. Pre-processing approaches modify the data distribution to satisfy fairness constraints, including data massaging \cite{kamiran2009classifying}, feature repair \cite{feldman2015certifying}, convex probabilistic transformations \cite{calmon2017optimized}, and maximum entropy formulations \cite{celis2019fair}. Optimal transport methods have been proposed to reweigh or shift distributions while minimizing divergence~\cite{gordaliza2019obtaining }. 
Post processing techniques instead modify model outputs to enforce parity constraints, for example through linear programming adjustments \cite{hardt2016equality} or Wasserstein barycentric projections \cite{pmlr-v115-jiang20a}. 
Although originally designed for mitigation, these techniques can also be repurposed to simulate strategic manipulation.

From the auditing perspective, recent work emphasizes the importance of access models and statistical testing. Discussions on AI auditing argue that, at minimum, black box query access is required for meaningful oversight and highlight the parallels between auditing and hypothesis testing, including burden of proof and error trade offs \cite{10.1145/3689904.3694711}. Statistical inference methods have been proposed to provide confidence bounds for fairness metrics using bootstrap guarantees \cite{JMLR:v25:23-0739}, and to adapt fairness testing to small or intersectional groups via size adaptive procedures. Complementary work studies black box audits for detecting group distribution shifts between training and deployment data using model queries \cite{juarez2022blackboxauditsgroupdistribution}, emphasizing that white box access is often unrealistic in practice.

In summary, fairness auditing is undergoing an arms race between auditees’ capacity to fake compliance and auditors’ ability to detect manipulation. Our work differs from prior fair washing studies by providing an explicit distributional projection formulation of audit evasion using entropic and optimal transport constructions, and by systematically analyzing the detectability of such manipulations through distribution based statistical tests. This positions fairness auditing as a constrained statistical inference problem, bridging distributional robustness, optimal transport, and hypothesis testing.

\section{Fairwashing detection by the supervisory authority: Statistical tests based on distributional distances} \label{sec:stat_tests}

We outline below potential strategies a supervisory authority could adopt to assess whether the auditee provided a sample drawn from the original data distribution to the compliance evaluation. We note $Q_{t}$ the distribution created by the auditee using $Q_n$, with $DI(f, Q_{t}, S) \geq t$. The non-malicious auditee always chooses $Q_{t} = Q_n$. Note that this distribution is never shared nor observable, even by the supervisory authority. The auditee presents a sample $\mathcal{D}_m \sim Q_t^{\bigotimes m}$, we denote by $\widehat{Q}_m$ the empirical distribution induced by $\mathcal{D}_m$ and let $t':= DI(f, \widehat{Q}_m,S)$. 

To verify the authenticity of this sample, the authority must be granted access to the full dataset upon request. This access enables the authority to reconstruct the ground-truth distribution $Q_n$ and determine whether the submitted data $\mathcal{D}_m$ arises from fairwashing manipulation $Q_t$ or follows the initial distribution $Q_n$. To assess representativeness, the authority must rely on statistical testing. Two main categories of tests are available.

The first includes hypothesis tests that evaluate, at a chosen confidence level, whether the empirical distribution $\widehat{Q}_{m}$ induced by the submitted sample is statistically similar to $Q_n$. In their study, \cite{fukuchi2020faking} apply a Kolmogorov–Smirnov (KS) test for one-dimensional data ($X \in \mathbb{R}^1$), and a test based on the Wasserstein distance for higher-dimensional settings ($X \in \mathbb{R}^k$, with $k > 1$). In our framework, we apply both the KS test and the Wasserstein test on the conditional distribution $\widehat{Y} \mid S$. 

The second approach evaluates whether the sample $\mathcal{D}_{m}$ could plausibly result from a random draw from the original distribution $Q_n$, by measuring a divergence or distance metric $d$. Let $Q_m^{(1)},\cdots,Q_m^{(B)}$ denote empirical distributions obtained from independent samples of size m drawn from $Q_n$. The idea is to test whether the observed distance value $d(\widehat{Q}_{m}, Q_n)$ lies within the acceptance region induced by the $(1-\alpha)$-quantiles of the reference distribution. The confidence interval is estimated from the empirical distribution of 
\begin{equation}
    d(Q_m^{(b)},Q_n), \text{\hspace{0.2cm}} b = 1,\cdots,B.
\end{equation}
For the distance metric $d$, we considered several options, including the Maximum Mean Discrepancy (MMD)~\cite{JMLR:v13:gretton12a}, the Wasserstein distance, and the Kullback–Leibler divergence. 
For the statistical tests we introduce based on distributional distances (all but KS), the detection threshold is based on the sample size provided by the auditee, making the results more robust to type I and II errors. All hypothesis tests reported in this paper use a 95\% confidence level.

\begin{remark} \textbf{Application to non-tabular data.}
The statistical tests (and later the methods we develop) are originally meant to handle tabular data. However they are applicable directly on images or text flattened as vectors. Yet, using the Wasserstein distance, based on the $L^2$ distance between individuals, might not be the best way to capture semantic similarity between images or token distributions. A way to circumvent this issue is to represent the images in another space, where the regular distances would have semantic meanings. The construction of such a space has already seen numerous works using PCA projections or latent spaces of AE, VAE or CNN classifiers. We present such results on the CelebA dataset \cite{liu2015faceattributes} in Section~\ref{app:sec:extension} of the Appendix.
\end{remark}

\section{Malicious auditee strategy: optimized fairwashing methods}
\label{sec:methods}
In the following, we will consider two distances: in Section~\ref{sec:KL}, one is related to the similarity for probabilistic inference (KL information), while in Section~\ref{sec:W2}, the other distance captures geometric information between distances (Monge-Kantorovitch a.k.a. Wasserstein distance). Consequently, fairwashing amounts to modifying the initial distribution of the data by providing a fake but plausible distribution $Q_t$ in order to achieve that $DI(f,Q_t,S) = t$ or $DI(f,Q_t,S) \geq t$. 

Developing such methods is unavoidable: only by simulating what a malicious auditee could do, can we study this scenario in order to raise regulators’ awareness, and evaluate which detection strategies are effective.

\subsection{Using entropic projection to fake fairness} \label{sec:KL}
We first present the main tool we are going to use to mimic how an auditee can build a fair-washed sample. To formalize the ideas presented in \ref{s:method}, we present a theorem on how to construct a distribution minimizing the KL divergence w.r.t to the true observation while satisfying a constraint on its mean, similar to a global fairness measure.

The KL information is $\KL ( P \Vert Q )=\int_{E}\log\frac{\dd P}{\dd Q}\dd P$, if $P\ll Q$ and
$\log\frac{\dd P}{\dd Q}\in L^1(P)$, and $+\infty$ otherwise. For any resulting dimension $k \geq 1$, let
$\Phi: Z = (X,S, \widehat{Y},Y) \in E \mapsto \Phi(X,S,\widehat{Y},Y) \in \mathbb{R}^k$
be a measurable function representing the magnitude of the stress deformation on the whole input.
Note that our results are stated for a generic function $\Phi$ of all variables $Z=(X,S,\widehat{Y},Y)$. This includes the case of functions depending only on $X$, $(X,Y)$ or $(X,\widehat{Y})$. 
We set for two vectors $x,y \in \mathbb{R}^k$ the scalar product as $\langle x,y \rangle=x^\top y$. The problem can be stated as follows: given the distribution $Q_n$, our aim is to construct a distribution close to $Q_n$ while satisfying a prescribed moment constraint, namely that the expectation of the chosen function $\Phi$ under the distribution equals a given value $t \in \mathbb{R}^k$. More precisely, a new distribution $Q_{t}$ satisfying the constraint $\int_{E} \Phi(z) \dd Q_t(z) = t$ and being the closest possible to the initial empirical distribution $Q_n$ in the sense of KL divergence, i.e. with $\KL ( Q_t \Vert Q_n )$ as small as possible. The following theorem, whose proof can be found in \cite{gems}, characterizes the distribution solution $Q_t$. 
\begin{theorem}[Entropic Projection under constraint] \label{thm:discrete:reweigth:multidim}
	Let $t \in \mathbb{R}^k$ and $\Phi : E \to \mathbb{R}^k$ be measurable.
	Assume that $t$ can be written as a convex combination of $\Phi(X_1,S_1, \widehat{Y}_1,Y_1) , \ldots , \Phi(X_n,S_n, \widehat{Y}_n,Y_n)$, with positive weights. Assume also that the empirical covariance matrix of $\Phi(X,S,\hat{Y}, Y)$ is invertible.

	Let $\mathcal{D}_{\Phi,t}$ be the set of all probability measures $P$ on $E$ such that $
	\int_{E} \Phi(x) \dd P(x)=t$.
		Then, \begin{equation} \label{thesol:Qn}
Q_t:=\mathrm{arginf}_{P\in\mathcal{D}_{\Phi,t}}\KL(P \Vert Q_n)
\end{equation}
    exists and is unique. It can also be computed as 

        \begin{equation}
            Q_t=
		\frac{1}{n}
		\sum_{i=1}^n
		\lambda_i^{(t)}
		\delta_{X_i , S_i, \widehat{Y}_i , Y_i}
        \end{equation}
        
        with, for $i\in\{1,\ldots,n\}$,
	$\lambda^{(t)}_i
		= \exp
		\bigg( \langle\xi(t) ,  \Phi(X_i,S_i, \widehat{Y_i},Y_i) \rangle - \log Z(\xi(t))
		\bigg)$, where for a  vector $\xi \in \mathbb{R}^k$, let $Z(\xi):=\frac{1}{n} \sum_{i=1}^n e^{\langle  \Phi(X_i,S_i, \widehat{Y}_i,Y_i) , \xi \rangle}$; and $\xi(t)$ is the unique minimizer of the strictly convex function
	$H(\xi):=\log Z(\xi)-\langle\xi,t\rangle$.
\end{theorem}
This theorem directly allows to fake statistical parity using entropic projection.
Let $t_{\text{init}}$ be such that $DI(f,Q_n,S) = t_{\text{init}}$. It enables to optimally (for the KL divergence) construct a distribution $Q_t$ such that $DI(f,Q_n,S) = t_{\text{new}} \geq t_{\text{init}}$ for a given $t_{\text{new}}$. The fairness improvement can be defined as $\Delta_{DI}:= DI(f,Q_t,S) - DI(f,Q_n,S)$.
Note that $ DI(f,Q_n,S)= \frac{\lambda_0 / n_0}{\lambda_1 / n_1}$ where for $i \in \{0,1\}$, $n_s = |\{i=1,\dots,n | S_i=s\}|$ and $\lambda_s = |\{i=1,\dots,n | \widehat{Y}_i=1 \land S_i=s\}|$. 
Note also that $\lambda_0 = \sum_{i=1}^n \widehat{Y}_i(1-S_i)$ and $\lambda_1 = \sum_{i=1}^n \widehat{Y}_iS_i$. Hence modifying the DI can be achieved applying Theorem~\ref{thm:discrete:reweigth:multidim} for $Z=(S,\widehat{Y})$ and selecting the function 
\begin{equation}
  \label{KLconstraint}
 \Phi(s, f(x)) = \begin{pmatrix} (1-s)f(x) \\ sf(x) \\ s \\ 1-s \end{pmatrix} \text{ and } t = \begin{pmatrix} \lambda_0 + \delta_0 \\ \lambda_1 - \delta_1 \\ n_1 \\ n_0 \end{pmatrix}. \end{equation}

Our purpose is to improve the perceived fairness of the model
. Accordingly, we only consider increasing the numerator $ + \delta_0 \geq 0$ and decreasing the denominator $- \delta_1 \leq 0$. Because $DI_\text{init} \leq DI_{\text{new}} \leq 1$, the new distribution will always be at least as fair as the original one.
\begin{proposition}[KL-fair washing method] \label{th:FWKL} Finding a solution $Q_t$ such that $\KL(Q_t \Vert Q_n)$ is minimum and $DI(f,Q_t,S)=DI(f,Q_n,S)+\Delta_{DI}$ is achieved by finding the solution to \eqref{thesol:Qn} with $\Phi$ defined as in \eqref{KLconstraint} and with the two possible choices of parameters:
\begin{itemize}
  \item Balanced case : set $\delta_0 = \delta_1$ and  $\delta_1 = \dfrac{\lambda_1}{1 + \frac{n_1}{n_0\Delta_{DI}}(1 + \frac{\lambda_0}{\lambda_1})}$
   
\item Proportional case : set $\dfrac{\delta_0}{n_0} = \dfrac{\delta_1}{n_1}$ and
    $ \delta_1 = \dfrac{\lambda_1}{1 + \frac{1}{\Delta_{DI}}\big(1 + \frac{n_1\lambda_0}{n_0\lambda_1}\big)} $
  \end{itemize}
  \end{proposition}

  \begin{remark}
        The balanced case corresponds to modifying 
        the individuals from both classes 
        equally, while the proportional one adjusts the amount of modification in proportion to the 
        classes sizes.   We refer respectively in the Experimental section to the two methods as \texttt{Entropic\_balanced} and \texttt{Entropic\_proportional}.
    \end{remark}

\subsection{Fairwashing using Optimal Transport.}
\label{sec:W2}
\subsubsection{Monge Kantorovich (MK) Projection} \label{sec:monge_kanto_proj} \label{sec:wass_grad}
For two distributions $P$ and $Q_n$ over $E \subset\R^d$ a compact subset, endowed with the norm $\|.\|$, recall that their 2 Monge-Kantorovich, a.k.a. Wasserstein distance, is defined as:
  \begin{equation}
   \label{eq:d_Wass}
  W_2^2(P,Q_n)= \min_{\pi \in \Pi(P,Q_n)} \int_{x\in E,y\in E} \parallel x-y \parallel^2 d\pi(x,y),
   \end{equation}
where $\Pi(P,Q_n)$ denotes the set of distributions on $E \times E$ with marginals $P$ and $Q_n$. We will write $T_{\sharp} Q = Q \circ T^{-1}$ to denote the push-forward of a measure by the transport map. As in Section~\ref{sec:KL}, consider for a given $k\geq1$, a continuous function $\Phi\colon E\to\R^k$
representing the constraints. For fixed $t\in\R^k$, the set $\mathcal{D}_{\Phi,t} = \{P\in\mathcal{M}(E)\mid \int_E \Phi(x)dP(x) = t\} $ is closed for the weak convergence and convex, since it is linear in $P$. The function $ P \mapsto W_2^2(P,Q_n)
$ is convex as it is the supremum of linear functionals by Kantorovich duality \cite{sant}, therefore the following projection problem $\mathrm{arginf}_{P\in\mathcal{D}_{\Phi,t}}W_2^2(P,Q_n)$ is well-defined.
   
\begin{theorem}\label{thm:empirical}
Consider $Q_n=\frac{1}{n} \sum_{i=1}^n \delta_{Z_i}$. Then $Q_t$ is a solution to 
\begin{equation}
\mathrm{arginf}_{P\in\mathcal{D}_{\Phi,t}}W_2^2(P,Q_n)
\end{equation}
if, and only if, it is defined as 
 $Q_{t}=T_{\lambda^\star}\sharp Q_n = \dfrac{1}{n} \sum_{i=1}^n \delta_{T_{\lambda^\star}(Z_i)}$,
where  $T_\lambda$ is defined as
\begin{equation}
 \label{eq:T_lambda}
 T_\lambda(Z_i) \in {\rm arg}\min_{x\in E} \|x-Z_i\|^2 - \langle \lambda , \Phi(x) \rangle
\end{equation}
  and  ${\lambda^\star}$ satisfies  $ t=\frac{1}{n}\sum_{i=n}^n \Phi(T_{\lambda^\star}(Z_i))$.
  \end{theorem}
\begin{remark}
\label{remark_ineq}
  The previous result stated for the empirical distribution is valid for any distribution $Q$. The constraint can be modified to include the condition $\mathcal{D}_{\Phi,t} = \{P\in\mathcal{M}(E)\mid \int_E \Phi(x)dP(x) \geq t\}$. This is detailed in Proposition~\ref{prop:extproje} in the Appendix.
\end{remark}

Following the framework of the previous section, for a fixed $\lambda$, we set $\Phi(x,s,f(x),y) = f(x)$. Then the constraint on the Disparate Impact $DI(f, Q, S) \geq t$, can be reformulated with double inequality given $S$.

Following Theorem~\ref{thm:empirical}, we compute for all $Z_i = x$, the solution $T_\lambda(x)$ of the minimization problem w.r.t $x$: 
\begin{equation}
  \label{eq:prblm_wass_impl}
  \mathcal{L}(x,\lambda) = \|Z_i - x\|_2^2 + \langle \lambda ,t- f(x)\rangle.
\end{equation}

This minimization does not have a closed form in general, but it can be achieved using a gradient descent using a learning step of $\eta$ and computing $x^k=x^{k-1}-\eta \frac{\partial \mathcal{L}}{\partial z}(x^k) $ with 
$\frac{\partial \mathcal{L}}{\partial x} = 2(z-x) - \langle \lambda,  \nabla_xf(x) \rangle$. We provide further implementation information in the Appendix, Sec.~\ref{app:sec:wass_grad}. 

Contrary to the entropic projection which is only a different sampling of the original individuals, this method creates new individuals. Yet it may create some out of bounds individuals by violation of some restriction of types (discrete variable staying discrete, i.e., $\text{age} = 1.002$) or of bounds ($\text{age} = -1$). Thereby, we created a variant of this method that constrains the covariates of new individuals: we transport, variable per variable, each covariate toward the nearest (for the $L_2$ norm) achieved value in the dataset; we call this variant the {1D-transport} variant. 

To summarize, we have introduced four methods: (1) \texttt{Grad\_balanced}, and (2) \texttt{Grad\_proportional}, which differ based on the gradient constraints satisfying $\delta_0 = \delta_1$ or $\frac{\delta_0}{n_0} = \frac{\delta_1}{n_1}$; and \\ (3) \texttt{Grad\_balanced\_1D-transport}, and (4) \texttt{Grad\_proportional\_1D-transport}, which apply the corresponding 1D-transport variant of each method.

\begin{figure}[t!]
   \centering
    \begin{minipage}{0.39\textwidth}
        \centering
        \scalebox{0.55}{ 


\begin{tikzpicture}[x=0.75pt,y=0.75pt,yscale=-1,xscale=1]

\draw  [draw opacity=0] (88.45,247.56) -- (289.11,247.56) -- (289.11,48.11) -- (88.45,48.11) -- cycle ; \draw   (88.45,247.56) -- (88.45,48.11)(268.45,247.56) -- (268.45,48.11) ; \draw   (88.45,247.56) -- (289.11,247.56)(88.45,67.56) -- (289.11,67.56) ; \draw    ;
\draw  [fill={rgb, 255:red, 191; green, 183; blue, 185 }  ,fill opacity=1 ] (76.45,247.58) .. controls (76.44,240.96) and (81.8,235.58) .. (88.42,235.56) .. controls (95.04,235.55) and (100.42,240.91) .. (100.44,247.53) .. controls (100.45,254.16) and (95.09,259.54) .. (88.47,259.55) .. controls (81.85,259.56) and (76.47,254.21) .. (76.45,247.58) -- cycle ;
\draw  [fill={rgb, 255:red, 0; green, 0; blue, 0 }  ,fill opacity=1 ] (256.45,247.58) .. controls (256.44,240.96) and (261.8,235.58) .. (268.42,235.56) .. controls (275.04,235.55) and (280.42,240.91) .. (280.44,247.53) .. controls (280.45,254.16) and (275.09,259.54) .. (268.47,259.55) .. controls (261.85,259.56) and (256.47,254.21) .. (256.45,247.58) -- cycle ;
\draw  [fill={rgb, 255:red, 0; green, 0; blue, 0 }  ,fill opacity=1 ] (76.45,67.58) .. controls (76.44,60.96) and (81.8,55.58) .. (88.42,55.56) .. controls (95.04,55.55) and (100.42,60.91) .. (100.44,67.53) .. controls (100.45,74.16) and (95.09,79.54) .. (88.47,79.55) .. controls (81.85,79.56) and (76.47,74.21) .. (76.45,67.58) -- cycle ;
\draw  [fill={rgb, 255:red, 191; green, 183; blue, 185 }  ,fill opacity=1 ] (256.45,67.58) .. controls (256.44,60.96) and (261.8,55.58) .. (268.42,55.56) .. controls (275.04,55.55) and (280.42,60.91) .. (280.44,67.53) .. controls (280.45,74.16) and (275.09,79.54) .. (268.47,79.55) .. controls (261.85,79.56) and (256.47,74.21) .. (256.45,67.58) -- cycle ;
\draw [line width=1.5]    (88.45,247.56) .. controls (67.98,211.11) and (63.79,127.45) .. (78,79.71) ;
\draw [shift={(78.67,77.56)}, rotate = 107.58] [color={rgb, 255:red, 0; green, 0; blue, 0 }  ][line width=1.5]    (19.89,-5.99) .. controls (12.65,-2.54) and (6.02,-0.55) .. (0,0) .. controls (6.02,0.55) and (12.65,2.54) .. (19.89,5.99)   ;
\draw [line width=0.75]  [dash pattern={on 4.5pt off 4.5pt}]  (268.45,67.56) .. controls (285.49,89.34) and (290.57,183.97) .. (279.99,238.9) ;
\draw [shift={(279.67,240.56)}, rotate = 281.38] [color={rgb, 255:red, 0; green, 0; blue, 0 }  ][line width=0.75]    (15.3,-4.61) .. controls (9.73,-1.96) and (4.63,-0.42) .. (0,0) .. controls (4.63,0.42) and (9.73,1.96) .. (15.3,4.61)   ;
\draw [line width=1.5]    (268.45,67.56) .. controls (237.14,43.92) and (160.03,35.8) .. (103.25,56.59) ;
\draw [shift={(100.67,57.56)}, rotate = 338.9] [color={rgb, 255:red, 0; green, 0; blue, 0 }  ][line width=1.5]    (19.89,-5.99) .. controls (12.65,-2.54) and (6.02,-0.55) .. (0,0) .. controls (6.02,0.55) and (12.65,2.54) .. (19.89,5.99)   ;
\draw [line width=1.5]    (88.45,247.56) .. controls (153.34,268.14) and (214.18,259.9) .. (258.01,253.92) ;
\draw [shift={(260.67,253.56)}, rotate = 172.23] [color={rgb, 255:red, 0; green, 0; blue, 0 }  ][line width=1.5]    (19.89,-5.99) .. controls (12.65,-2.54) and (6.02,-0.55) .. (0,0) .. controls (6.02,0.55) and (12.65,2.54) .. (19.89,5.99)   ;

\draw (191,263) node [anchor=north west][inner sep=0.75pt]   [align=left] {S : sensitive variable};
\draw (25,25) node [anchor=north west][inner sep=0.75pt]   [align=left] {Y : decision};
\draw (97,221) node [anchor=north west][inner sep=0.75pt]   [align=left] {(0,0)};
\draw (226,76) node [anchor=north west][inner sep=0.75pt]   [align=left] {(1,1)};
\draw (225,218) node [anchor=north west][inner sep=0.75pt]   [align=left] {(1,0)};
\draw (98,78) node [anchor=north west][inner sep=0.75pt]   [align=left] {(0,1)};

\end{tikzpicture}}
        \caption{Admissible modifications of $\tau : {0,1}^2 \mapsto {0,1}^2$ that increase Disparate Impact using the \texttt{Replace} method.}
        \label{fig:Alteration}
    \end{minipage}
    \hfill
\begin{minipage}{0.59\textwidth}
\centering 
\begin{algorithm}[H] 
\caption{\texttt{Replace($S$,$\widehat{Y}$)} algorithm} 
\label{alg:Replace} 
\begin{algorithmic}[1] 
\State $Z^j = (Z_1, \cdots, Z_n), Z_i = (S_i, \widehat{Y}_i), t\in ]0,1[$
\State $\tau_i := (\tau, j)$ such as $\tau \in \mathcal{A}, i\in 1, \cdots, n$ 
\While{DI$(Z^j) < t$} 
\State $\tau_{i_0} \in \text{argmax } \text{ DI}(\tau_{i_0}(Z^j)) - \text{DI}(Z^j)$ \\ with $\tau_{i_0}(Z^j) := (Z_i,\cdots, \tau(Z_{i_0}), \cdots, Z_n)$
\State $Z^{j} \leftarrow Z^{j+1} = \tau_{i_0}(Z^j)$
\EndWhile 
\State
\Return $Z^j$
\end{algorithmic}
\end{algorithm}
\end{minipage}
\vspace{-0.2cm}
\end{figure}
\subsubsection{Faking Statistical Parity using sensitive attributes replacement.} \label{sec:key_attr}

In the previous subsection, we assumed that the auditor could compute $f(X)$ from any observation $X$. Hereafter, we assume that it does not have access to the model $f$ and only requests the outcome of the algorithm $\widehat{Y}$, without computing it. 
 
In this setting, faking fairness can be achieved by manipulating only the outcomes and sensitive attributes associated with each individual. Let $Z_i=(S_i, \widehat{Y}_i)$ and $ Q_n=\frac{1}{n} \sum_{i=1}^n \delta_{S_i, \widehat{Y}_i}.$ In this context, a solution to the optimization problem can be achieved as follows. First, note that $\widehat{Y} \in \{ 0,1\}$ and $S \in \{ 0,1\} $, thus we only have 4 possible values for the points. Each individual with characteristic $Z_i\in \{0,1\}^2$ can be modified to the individual $\tau(Z_i)=(\tau_S(S_i),\tau_{\widehat{Y}}(\widehat{Y}_i))\in \{0,1\}^2$. Not all solutions improve the disparate impact and we can therefore restrict ourselves to a set of admissible changes $\tau \in \mathcal{A}$ as pointed in Fig.~\ref{fig:Alteration}, see Section~\ref{app:sec:wass_sampling} in the Appendix. We approximate the exact solution by an iterative method starting from $Z = (Z_1, \cdots, Z_n)$ and testing every possible modification $Z^j = (Z_1, \cdots, Z_n)$ maximizing the $DI$ at each step $j$. The method based on this algorithm is denoted by \texttt{Replace($S$,$\widehat{Y}$)} in our experiments (see Alg.~\ref{alg:Replace}). 

\textbf{Faking Statistical Parity using constrained matching.} \label{sec:wass_matching} In the previous case, only the labels are transported while the observations $X_i$ are not taken into account. A natural variant consists in combining previous minimization scheme and adding a discrete displacement on the variables $X$. Namely, we define a matching algorithm using $ Z=(X,S,\widehat{Y})$ and $\tau(Z_i)= Z_k, \text{ with } k \in \{ 1,\cdots, n \}.$ We use the same procedure as Alg.~\ref{alg:Replace} with the newly defined $\tau$, but at each iteration $j$ of the while loop, we maximize for every candidate $\tau_{i_0}$: 
\begin{equation}
    \dfrac{\text{DI}(\tau_{i_0}(Z^j)) - \text{DI}((Z^j))}{\|\tau_{i_0}(Z^j) - Z^j\|}
\end{equation}
In our experiments, we refer to the method based on this algorithm as $M_{W(X,S,\widehat{Y})}$. 

\begin{remark}
  This algorithm transports individuals towards others ($\tau(Z_i)= Z_k$), therefore, contrary to its counterpart, it can be used in any type of audit (with or without access to the model). 
\end{remark} 

\begin{table}[t!]
 \centering
  \caption{Dataset presentation, protected variable (S) associated, and original Disparate Impact (DI)}

 \begin{tabular}{l lll lll l}
  \toprule
  & Adult & INC & TRA & MOB & BAF & EMP & PUC  \\
  \midrule
  S chosen & Sex & Sex & Sex & Age & Age & Disability & Disability \\
  DI & 0.30 & 0.67 & 0.69 & 0.45 & 0.35 & 0.30 & 0.32 \\
  \bottomrule
 \end{tabular}

  \label{tab:Dataset}
\end{table}

\begin{figure}[tb!]
\centering
  \begin{minipage}{0.41\textwidth}
    \centering
    \includegraphics[width=1\textwidth, height=5cm]{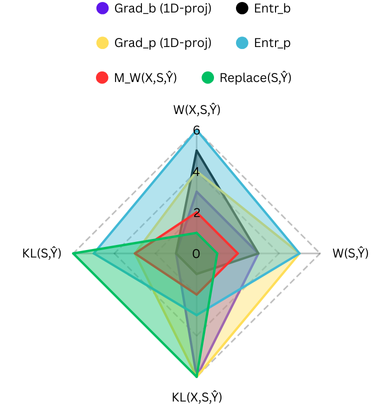}
    \caption{Radar graph ranking $\displaystyle \KL$ and $W$ similarity results depending on the fairwashing method (lower is better). The visualization highlights why $M_{W(X,S,\widehat{Y})}$, considering $X$ or not on the similarity provides the most balanced overall performance. Entr\_b and Entr\_p are respectively for \texttt{Entropic\_balanced} and \texttt{Entropic\_proportional}. 
    }
    \label{fig:RadarChart}
  \end{minipage}
  \hfill
  \begin{minipage}{0.53\textwidth}
    \centering
\includegraphics[width=1\textwidth, height=5cm]{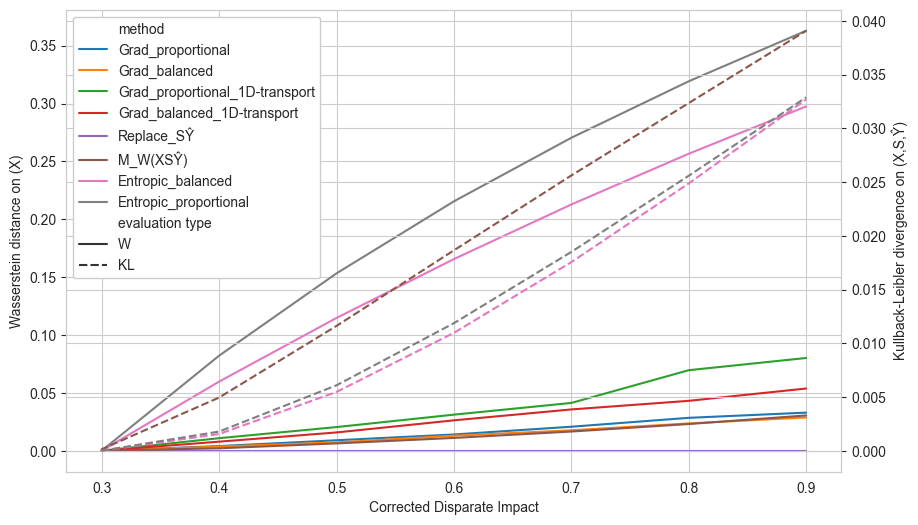}
\caption{
Line plot illustrating the trade-off between fairness correction and distribution shift (on $X$ for the Wasserstein distance and $(X, S, \widehat{Y})$ for the $\KL$) with the Adult dataset. KL divergence and Wasserstein distance, between the fully modified and the original datasets are reported for each method. Methods with infinite $\KL$ are omitted. 
}
\label{fig:Unbiasing_Adult}
  \end{minipage}
\end{figure}

\section{Experimental settings}
\label{sec:expe}


\textbf{Datasets.} 
We use 7 benchmarking datasets : Adult Census Income dataset 
where $Y$ is whether an individual's income is above 50k (Adult) \cite{adult_2}. We also use 5 benchmark datasets \cite{ding2021retiring} which records information about the USA's population, including income (INC), mobility (MOB), employment (EMP), travel time to work (TRA) and public system coverage (PUC). We also include the Bank Account Fraud generated dataset (BAF) \cite{jesus2022turningtablesbiasedimbalanced}. We refer to Table \ref{tab:Dataset} for the sensitive variable and the original Disparate Impact (DI) of each dataset. In our experiments, a part of the dataset was used to learn the outcome decision, we selected 5k individual of the test set for Adult, and 20k individuals of the test set for the other datasets.

\textbf{Neural network predictions.} For every dataset classification task, we trained an MLP. As we are working only with tabular data, we provide a $\widehat{Y}$ with a multilayer perceptron (MLP) neural network $f$ ending with a sigmoid activation function ($f(x)\in[0,1]$). Accuracy performance was not the main goal in this study, see in the Appendix Sec.~\ref{app:sec:accuracy}. For that reason, the model's architecture and the training protocol are classical : 80\% of the dataset size was used as training set (including validation set) and the remainaing 20\% for the test set. We set the starting learning rate at $0.001$ using the AdamW Schedule-free optimizer \cite{defazio2024road}. All configurations for each dataset are available in our \href{https://github.com/ValentinLafargue/Inspection}{GitHub repository}. The models have been uploaded to \href{https://huggingface.co/ValentinLAFARGUE/EIF-biased-classifiers}{Hugging Face}, and the original and manipulated distributions are available at \href{https://huggingface.co/datasets/ValentinLAFARGUE/EIF-Manipulated-distributions}{Hugging Face Datasets}.

\section{Results} 


\textbf{Fairness cost: distribution shift per method.} Fig~\ref{fig:RadarChart} illustrates the comparative performance of each method across different distance metrics ($\displaystyle \KL$ , $W $). 
Specifically, $\displaystyle \KL$ and $W$ quantify the magnitude of the distributional shift, either marginally or conditional on the input features $X$, and thus assess each method’s detectability by statistical tests.
We also provide complementary results on simulated data and the computation time and memory cost of each method in the Appendix (see Sec~\ref{app:sec:simu}).
The smallest surface area in the radar chart is archived by \texttt{$M_{W(X,S,\widehat{Y})}$}, hence, given the results, this method appears to be the most suitable method for someone seeking to disguise their dataset, as it significantly improves the DI while preserving a distribution close to the original data. 

\begin{table*}[tb!]
 \centering
 \caption{Results of the 7 tests independently; for each unbiasing method (DI=0.8) and datasets. Sampling is stopped as soon as one sample do not reject the $\mathcal{H}_0$ hypothesis, or after 30 tries if none do.
 The symbol $-$ means the method was undetected by the tests for both sampling sizes of 10\% and 20\% ($\mathcal{H}_0$ not rejected), 
 \scalebox{1.5}{$\iscircle$} means that only the 20\% sampling size was undetected ($\mathcal{H}_0$ not rejected for 20\% and rejected for 10\%); 
 and $\circledcirc$ means that the method was detected at both 10\% and 20\% sample sizes ($\mathcal{H}_0$ rejected). Positional and color coding indicate which test each result corresponds to, in the following order and color scheme: \(\color{red} \KL(X, S, \widehat{Y})\), \(\color{orange} \KL(S, \widehat{Y})\), \(\color{green} W(X, S, \widehat{Y})\), 
\(\color{blue} W(S, \widehat{Y})\), \(\color{purple} \text{K-S}(\widehat{Y})\), \(\color{brown} \text{MMD}(X, S, \widehat{Y})\), \(\text{MMD}(S, \widehat{Y})\) . \texttt{Grad\_proportional} (Grad\_p) and \texttt{Grad\_balanced} (Grad\_b) have been merged with their 1D counterpart due to identical test results.  Entropic\_b and Entropic\_p are respectively for \texttt{Entropic\_balanced} and \texttt{Entropic\_proportional}. }
 \scalebox{0.75}{\begin{tabular}{l cc cc cc}
  \toprule
    & \multicolumn{5}{c}{Methods}                   \\ \cmidrule(r){2-7} \\
    Dataset     &  Grad\_p(1D-t)& Grad\_b(1D-t) & Rep $(S,\widehat{Y})$ & $M_{W(X,S,\widehat{Y})}$ & Entropic\_b & Entropic\_p \\
    \midrule
ADULT   & $\color{red}  \circledcirc$ $\color{orange}  \circledcirc$ $ \color {green} -$ $\color{blue}  \circledcirc$ $\color{purple}  \circledcirc$ $\color{brown}-$ $\circledcirc$
        & $\color{red}  \circledcirc$ $\color{orange}  \circledcirc$ $ \color{green} -$ $\color{blue}  \circledcirc$ $\color{purple}  \circledcirc$ $\color{brown}-$ $\circledcirc$
        & $\color{red}  \circledcirc$ $\color{orange}  \circledcirc$ $ \color{green} -$ $\color{blue}  \circledcirc$ $\color{purple}  \circledcirc$ $\color{brown}-$ $\circledcirc$
        & \scalebox{1.5}{$\color{red}\iscircle$} $\color{orange}  \circledcirc$ $ \color{green} -$ $\color{blue}  \circledcirc$ $\color{purple}  \circledcirc$ \scalebox{1.5}{$\color{brown} \iscircle$} $\circledcirc$
        & $ \color{red} -$ $\color{orange}  \circledcirc$ $ \color{green} -$ $\color{blue}  \circledcirc$ $\color{purple}  \circledcirc$ $\color{brown}-$ $\circledcirc$
        & $ \color{red} -$ $\color{orange}  \circledcirc$ $ \color{green} -$ $\color{blue}  \circledcirc$ $\color{purple}  \circledcirc$ $\color{brown}-$ $\circledcirc$\\
        
EMP     & $\color{red}  \circledcirc$ $\color{orange}  \circledcirc$ $\color{green}  \circledcirc$ $\color{blue}  \circledcirc$ $\color{purple}  \circledcirc$ $\color{brown}\circledcirc$ $\circledcirc$ 
        & $\color{red}  \circledcirc$ $\color{orange}  \circledcirc$ \scalebox{1.5}{$\color{green} \iscircle$} $\color{blue}  \circledcirc$ $\color{purple}  \circledcirc$ \scalebox{1.5}{$\color{brown} \iscircle$} $\circledcirc$
        & $\color{red}  \circledcirc$ $\color{orange}  \circledcirc$ $ \color{green} -$ $\color{blue}  \circledcirc$ $\color{purple}  \circledcirc$ $\color{brown}-$ $\circledcirc$
        & $\color{red}  \circledcirc$ $\color{orange}  \circledcirc$ $ \color{green} -$ $\color{blue}  \circledcirc$ $\color{purple}  \circledcirc$ \scalebox{1.5}{$\color{brown} \iscircle$} $\circledcirc$
        & \scalebox{1.5}{$\color{red}\iscircle$} $\color{orange}  \circledcirc$ $\color{green}  \circledcirc$ $\color{blue}  \circledcirc$ $\color{purple}  \circledcirc$ \scalebox{1.5}{$\color{brown} \iscircle$} $\circledcirc$
        & \scalebox{1.5}{$\color{red}\iscircle$} $\color{orange}  \circledcirc$ $\color{green}  \circledcirc$ $\color{blue}  \circledcirc$ $\color{purple}  \circledcirc$ $\color{brown} -$ $\circledcirc$\\
        
INC     & $\color{red}  \circledcirc$ $ \color{orange} -$ $ \color{green} -$ $ \color{blue} -$ $ \color{purple} -$ $\color{brown}-$ $-$
        & $\color{red}  \circledcirc$ $ \color{orange} -$ $ \color{green} -$ $ \color{blue} -$ $ \color{purple} -$ $\color{brown}-$ $-$
        & $\color{red}  \circledcirc$ $ \color{orange} -$ $ \color{green} -$ $ \color{blue} -$ $ \color{purple} -$ $\color{brown}-$ $-$
        & $ \color{red} -$ $ \color{orange} -$ $ \color{green} -$ $ \color{blue} -$ $ \color{purple} -$ $\color{brown}-$ $-$
        & $ \color{red} -$ $ \color{orange} -$ $ \color{green} -$ $ \color{blue} -$ $ \color{purple} -$ $\color{brown}-$ $-$
        & $ \color{red} -$ $ \color{orange} -$ $ \color{green} -$ $ \color{blue} -$ $ \color{purple} -$ $\color{brown}-$ $-$\\
        
MOB     & $\color{red}  \circledcirc$ $\color{orange}  \circledcirc$ $\color{green}  \circledcirc$ $\color{blue}  \circledcirc$ $\color{purple}  \circledcirc$ $\color{brown} \circledcirc$ $\circledcirc$
        & $\color{red}  \circledcirc$ $\color{orange}  \circledcirc$ $ \color{green} -$ $\color{blue}  \circledcirc$ $\color{purple}  \circledcirc$ $\color{brown} -$ $\circledcirc$
        & $\color{red}  \circledcirc$ $\color{orange}  \circledcirc$ $ \color{green} -$ $\color{blue}  \circledcirc$ $\color{purple}  \circledcirc$ $\color{brown}-$ $\circledcirc$
        & $\color{red}  \circledcirc$ $\color{orange}  \circledcirc$ $ \color{green} -$ $\color{blue}  \circledcirc$ $\color{purple}  \circledcirc$ \scalebox{1.5}{$\color{brown} \iscircle$} $\circledcirc$
        & \scalebox{1.5}{$\color{red}\iscircle$} $\color{orange}  \circledcirc$ \scalebox{1.5}{$\color{green}\iscircle$} $\color{blue}  \circledcirc$ $\color{purple}  \circledcirc$   \scalebox{1.5}{$\color{brown} \iscircle$} $\circledcirc$
        & \scalebox{1.5}{$\color{red}\iscircle$} $\color{orange}  \circledcirc$ $\color{green}  \circledcirc$ $\color{blue}  \circledcirc$ $\color{purple}  \circledcirc$  $\color{brown} \circledcirc$ $\circledcirc$\\
        
PUC     & $\color{red}  \circledcirc$ $\color{orange}  \circledcirc$ $\color{green}  \circledcirc$ $\color{blue}  \circledcirc$ $\color{purple}  \circledcirc$  $\color{brown} \circledcirc$ $\circledcirc$
        & $\color{red}  \circledcirc$ $\color{orange}  \circledcirc$ \scalebox{1.5}{$\color{green}\iscircle$} $\color{blue}  \circledcirc$ $\color{purple}  \circledcirc$  \scalebox{1.5}{$\color{brown} \iscircle$} $\circledcirc$
        & $\color{red}  \circledcirc$ $\color{orange}  \circledcirc$ \scalebox{1.5}{$\color{green}\iscircle$} $\color{blue}  \circledcirc$ $\color{purple}  \circledcirc$ \scalebox{1.5}{$\color{brown} \iscircle$} $\circledcirc$
        & $\color{red}  \circledcirc$ $\color{orange}  \circledcirc$ \scalebox{1.5}{$\color{green}\iscircle$} $\color{blue}  \circledcirc$ $\color{purple}  \circledcirc$ $\color{brown} \circledcirc$ $\circledcirc$
        & \scalebox{1.5}{$\color{red}\iscircle$} $\color{orange}  \circledcirc$ $\color{green}  \circledcirc$ $\color{blue}  \circledcirc$ $\color{purple}  \circledcirc$ \scalebox{1.5}{$\color{brown} \iscircle$} $\circledcirc$
        & \scalebox{1.5}{$\color{red}\iscircle$} $\color{orange}  \circledcirc$ $\color{green}  \circledcirc$ $\color{blue}  \circledcirc$ $\color{purple}  \circledcirc$ $\color{brown} \circledcirc$ $\circledcirc$ \\
        
TRA     & $\color{red}  \circledcirc$ $ \color{orange} -$ $ \color{green} -$ $ \color{blue} -$ $ \color{purple} -$ $\color{brown}-$ $-$
        & $\color{red}  \circledcirc$ $ \color{orange} -$ $ \color{green} -$ $ \color{blue} -$ $ \color{purple} -$ $\color{brown}-$ $-$
        & $\color{red}  \circledcirc$ \scalebox{1.5}{$\color{orange}\iscircle$} $ \color{green} -$ $ \color{blue} -$ $ \color{purple} -$ $\color{brown}-$ $-$
        & $ \color{red} -$ $ \color{orange} -$ $ \color{green} -$ $ \color{blue} -$ $ \color{purple} -$ $\color{brown}-$ $-$
        & $ \color{red} -$ $ \color{orange} -$ $ \color{green} -$ $ \color{blue} -$ $ \color{purple} -$ $\color{brown}-$ $-$
        & $ \color{red} -$ $ \color{orange} -$ $ \color{green} -$ $ \color{blue} -$ $ \color{purple} -$ $\color{brown}-$ $-$\\
        
BAF     & $\color{red}  \circledcirc$ \scalebox{1.5}{$\color{orange}\iscircle$} $ \color{green} -$ $ \color{blue} -$ $ \color{purple} -$ $\color{brown}-$ $-$
        & $\color{red}  \circledcirc$ \scalebox{1.5}{$\color{orange}\iscircle$} $ \color{green} -$ $ \color{blue} -$ $ \color{purple} -$ $\color{brown}-$ $-$
        & $\color{red}  \circledcirc$ $ \color{orange} -$ $ \color{green} -$ $ \color{blue} -$ $ \color{purple} -$ $\color{brown}-$ $-$
        & $ \color{red} -$ $ \color{orange} -$ $ \color{green} -$ $ \color{blue} -$ $ \color{purple} -$ $\color{brown}-$ $-$
        & $ \color{red} -$ $ \color{orange} -$ $ \color{green} -$ $ \color{blue} -$ $ \color{purple} -$ $\color{brown}-$ $-$
        & $ \color{red} -$ \scalebox{1.5}{$\color{orange}\iscircle$} $ \color{green} -$ $ \color{blue} -$ $ \color{purple} -$ $\color{brown}-$ $-$\\
    \bottomrule
    \vspace{-0.7cm}
  \end{tabular}
  
  }
  
  \label{tab:test_individual_result}
\end{table*}

\textbf{Fraud detection through distributional shifts.} Using statistical tests, is it possible to detect a fraud attempt based on the provided sample $\mathcal{D}_{m}$ and the original dataset $Q_n$? For each optimized fairwashing method presented in Sec.~\ref{sec:methods}, we obtain a modified distribution $Q_t$ for every dataset. For each method-dataset combination, we draw multiple samples $\mathcal{D}_{m}$ such that $DI(f, \widehat{Q}_{m}, S) \geq 0.8$, where $\widehat{Q}_{m}$ denotes the empirical distribution induced by the sample. More precisely, we repeatedly sample from each $Q_t$ and apply the seven statistical tests described in Section~\ref{sec:stat_tests}. For every combination of fairwashing method, dataset, and statistical test, we record whether the null hypothesis $\mathcal{H}_0$ is rejected, thereby assessing whether the original and sampled distributions can be considered equivalent. Table~\ref{tab:test_individual_result} reports the results obtained from samples drawn from the modified distributions for each dataset. The sample sizes considered are 10\% and 20\% of the original dataset (i.e., $m = \frac{n}{10}$ and $m = \frac{n}{5}$, respectively). Rejection of the null hypothesis $\mathcal{H}_0$ indicates that the fairwashing strategy employed by a malicious auditee would be detected by the supervisory authority. Consequently, a malicious auditee would seek to construct a fair yet detection-free sample by repeatedly drawing samples from $Q_t$. Additional details regarding the probability of passing the statistical tests are provided in Section~\ref{app:sec:tries} of the Appendix.

Methods modifying individual characteristics (\texttt{Grad} methods) are easily detected (rejection of \( \mathcal{H}_0 \)) regardless of the sampling size. 
The fairwashing done by the $M_{W(X,S,\widehat{Y})}$ and Entropic-based methods is undetected 
for the INC, TRA and BAF datasets. 
For the TRA and INC datasets, the DIs of the original data were close to that of the modified data (see Table~\ref{tab:highest_undetected_unbiasing}), implying that the required modifications were smaller than the others, and therefore the fairwashing was more difficult to detect. For the BAF dataset, because $\mathbb{E}(Y) \approx 0.01$, only limited individual modifications were needed. The results on the BAF dataset shows that the original Disparate Impact value is not the only criteria to consider when evaluating fairwashing detection difficulty. We provide complementary information in the Appendix about the simulation setup in Sec~\ref{app:sec:simu}, which confirms the superior trade-off achieved by $M_{W(X,S,\widehat{Y})}$ and we report the computational costs of the different methods in Section~\ref{app:sec:cost} in the Appendix.

\textbf{Trade-off: DI improvement vs distribution shift.} \label{sec:opti_result}
Fig.~\ref{fig:Unbiasing_Adult} illustrates the trade-off between fairness correction and distribution shift on the Adult datasets by the Wasserstein distance and KL divergence between the full original and modified distributions. \texttt{Replace $(S,\widehat{Y})$}, \texttt{$M_{W(X,S,\widehat{Y})}$} and \texttt{Grad} variant methods preserve the structure of the input space and are better alternatives to the entropic projection method. 
We recall that since \texttt{Replace} only modifies $S$ and $\widehat{Y}$, it naturally leads to null difference between distributions on $X$. Note that the relative ordering of methods presented on the Adult dataset remains largely consistent across datasets.

\begin{table*}[tb!]
    \centering
    \caption{Highest undetected achievable Disparate Impact for each dataset, sample size (Size) and fairwashing method. The symbol -- indicates that some methods couldn't reach a DI improvement. To emphasize the best method to use in order to deceive the auditor, we put the DI achieved in bold when one or two overperformed the others.}
    \scalebox{0.75}{
    \begin{tabular}{ll c cccc cccc}
\toprule
Dataset & Original & Size (\%) & Grad\_p & Grad\_b & Grad\_p 1D & Grad\_b 1D & Rep $(S, \widehat{Y})$ & Entr\_b & Entr\_p & $M_{W(X,S,\widehat{Y})}$ \\
\midrule
ADULT & 0.30 & 10 & 0.47 & 0.43 & 0.49 & 0.44   & 0.50 & \textbf{0.54} & 0.52 & \textbf{0.54} \\
      &      & 20 & 0.39 & 0.40 & 0.38 & 0.39   & 0.41 & \textbf{0.42} & 0.41 & \textbf{0.42} \\
EMP   & 0.30 & 10 & --   & --   & --   & --     & --   & 0.36          & 0.36 & \textbf{0.37} \\
      &      & 20 & --   & --   & --   & --     & --   & 0.34          & \textbf{0.36} & 0.35 \\
INC   & 0.67 & 10 & 0.75 & --   & --   & --     & 0.88 & 0.94          & \textbf{0.95} & 0.93 \\
      &      & 20 & --   & --   & --   & --     & 0.83 & 0.83          & \textbf{0.84} & \textbf{0.84} \\
MOB   & 0.45 & 10 & 0.53 & 0.51 & --   & 0.50   & \textbf{0.53}        & 0.52          & -- & 0.52 \\
      &      & 20 & --   & --   & --   & 0.48   & 0.50 & 0.50          & --   & 0.50 \\
PUC   & 0.32 & 10 & --   & --   & --   & --     & --   & 0.33          & \textbf{0.35} & \textbf{0.35} \\
      &      & 20 & --   & --   & --   & --     & --   & --            & --   & --   \\
TRA   & 0.69 & 10 & 0.72 & 0.79 & 0.77 & 0.73   & 0.80 & 0.83          & 0.84 & \textbf{0.85} \\
      &      & 20 & --   & --   & --   & --     & 0.77 & 0.79          & 0.79 & \textbf{0.80} \\
BAF   & 0.35 & 10 & --   & --   & --   & --     & --   & 1             & 1    & 1    \\
      &      & 20 & --   & --   & --   & --     & --   & 0.77          & \textbf{0.80} & 0.79 \\
    \bottomrule
    \vspace{-0.7cm}
    \end{tabular}
    }
    \label{tab:highest_undetected_unbiasing}
\end{table*}

\textbf{Fairest undetected sample.} Given the results reported in Table~\ref{tab:test_individual_result}, which show that achieving $DI(f, \widehat{Q}_m, S) \geq 0.8$ while evading detection is generally not possible in multiple datasets, we investigate the maximum fairness improvement that a malicious auditee could achieve without being detected by the supervisory authority. This analysis follows the fairness-detectability trade-off discussed previously. Note that such an optimization would only be possible if the malicious auditee were aware of the statistical tests employed by the supervisory authority, which constitutes a reasonable legal assumption.

Table~\ref{tab:highest_undetected_unbiasing} reports the highest Disparate Impact (DI) values achieved by samples that were not rejected by any of the seven statistical tests. For each modified distribution $Q_t$, we generate 100 random samples comprising either $10\%$ or $20\%$ of the original dataset size. We then progressively decrease the target fairness level $t$ in increments of $0.05$. More precisely, the procedure starts from $Q_{t=1}$. If at least one undetected sample $\mathcal{D}_m$ satisfying $DI(\mathcal{D}_m) \geq t$ is found, we retain the sample with the highest observed DI value, denoted $t'$. Otherwise, the procedure is repeated with a lower target fairness level, setting $t \leftarrow t - 0.05$. The process continues until $t - 0.05 < DI(Q_n)$. The samples drawn from the four \texttt{Grad} method variants and the \texttt{Replace} method failed to increase DI without detection on three common datasets (EMP, PUC, and BAF). This outcome is primarily due to the consistent rejection of these samples by the $\KL$-based tests (100\% rejection rate). These results emphasize the importance of combining multiple statistical tests: methods optimized with respect to the Wasserstein distance are effectively detected by tests based on the $\KL$ divergence. This observation highlights the advantage of the \texttt{Entropic} methods and the matching approach $M_{W(X,S,\widehat{Y})}$, which avoid scenarios where $\KL(Q_t \Vert Q_n) = +\infty$ and therefore exhibit greater robustness. Overall, $M_{W(X,S,\widehat{Y})}$ consistently achieves the strongest performance. 

The results reported in Table~\ref{tab:highest_undetected_unbiasing} also underscore the influence of sample size. For example, on the PUC dataset, undetected fairwashing was possible when using samples comprising 10\% of the original dataset, but not when using 20\% samples. This finding suggests that, as the sample size increases relative to the original dataset, it becomes increasingly difficult to improve DI without being detected. We further investigate this phenomenon in Fig.~\ref{app:fig:sample_size} of Appendix~\ref{sec:futher_studies}.

\section{Concluding guidelines}

Our results highlight a key limitation of fairness auditing: global fairness metrics can be manipulated if the evaluation distribution is controlled. We showed both theoretically and empirically that samples can be constructed to appear representative while satisfying fairness constraints, even when the full dataset violates them. Consequently, if the auditee controls the audit sample, fairness compliance may be artificially created. Representativeness must therefore be treated as a primary objective of the audit process. Auditors should avoid letting the auditee freely select the audit subset and, whenever possible, retain the ability to access the full dataset or request additional samples to estimate the reference distribution.

We propose a practical strategy for assessing representativeness based on combining statistical tests that capture complementary properties of distributions. While this significantly reduces the room for manipulation, our experiments show that carefully designed transformations can still remain statistically undetectable, especially if the auditee is aware of the supervisory authority detection protocol. In practice, the most effective lever to limit manipulation is the audit sample size. Larger samples substantially reduce the space of undetectable distributional shifts. We therefore recommend requiring sufficiently large samples, combining multiple representativeness tests. We hope this work contributes to the development of more robust auditing frameworks and provides practical guidance for regulators and practitioners seeking to ensure that fairness claims remain meaningful under adversarial incentives.

\subsection*{Ethical Considerations}

This work studies how malicious actors could manipulate audit datasets to appear compliant with fairness metrics such as Disparate Impact. Our objective is to expose these vulnerabilities in order to strengthen auditing procedures and regulatory oversight. By analyzing both manipulation strategies and statistical detection methods, we aim to support the development of more robust fairness auditing frameworks.

\section*{Reproducibility statement}

The algorithms corresponding to each proposed fairwashing method are detailed in the paper. For the simplified versions of the \texttt{Replace} $(S,Y)$ and $M_{W(X,S,Y)}$ methods, we refer the reader to Alg.~\ref{alg:Replace} in the main paper. The full, non-simplified version is provided in Alg.~\ref{app:alg:DI_W_miti} in the Appendix. Additionally, the Wasserstein-based gradient optimization fairwashing methods are described in Alg.~\ref{app:alg:MK_c_proj}, also in the Appendix.

Our experiments on both our simulated dataset and publicly available datasets \cite{adult_2, ding2021retiring, jesus2022turningtablesbiasedimbalanced, liu2015faceattributes}, together with all code required to reproduce the results reported in this paper using fixed random seeds and archived intermediate results, are available through our \href{https://github.com/ValentinLafargue/Inspection}{GitHub repository}, \href{https://huggingface.co/ValentinLAFARGUE/EIF-biased-classifiers}{Hugging Face model repository}, and \href{https://huggingface.co/datasets/ValentinLAFARGUE/EIF-Manipulated-distributions}{Hugging Face dataset repository}.

Our GitHub repository is organized as follows:
\begin{itemize}
\item \textbf{Data}: datasets (primarily CSV files).
\item \textbf{Pre-processing}: Jupyter notebooks.
\item \textbf{Src}: Python functions implementing our fairwashing methods.
\item \textbf{Project}: network training and inference, fairness evaluation, fairwashing, and fraud detection using statistical tests.
\item \textbf{Results}: final and intermediate results (CSV, NPY, and JSON files).
\end{itemize}

Because GitHub imposes file size limits, we host the datasets on Google Drive and Hugging Face.

\begin{credits}
\subsubsection{\discintname}
The authors have no competing interests to declare that are
relevant to the content of this article.

\subsubsection{Fundings}
This paper has been partially funded by the Agence Nationale de la Recherche under grants ANR-23-CE23-0029 Regul-IA. The authors also acknowledge the support of the AI Cluster ANITI (ANR-23-IACL-0002).

\end{credits}

\newpage


\appendix

\chapter*{Appendix}
\startcontents[sections]
\printcontents[sections]{l}{0}{\setcounter{tocdepth}{2}}

\section{Other fairness metrics}
\label{app:sec:other_fairness_metric}

In our paper, we focused on the Disparate Impact (DI) fairness metric, as it is one of the most widely used metrics. While this choice is justified, it is natural to wonder whether our results are specific to this metric or whether they are metric-agnostic. 

Our fairwashing method could have been implemented to minimize the distribution shift while being constrained to other global fairness metrics as long as we can write them as an integrable function or a combination of integrable functions. This condition is not very restrictive in our case. In fact, it only excludes the individual fairness metric, whereas most global fairness metrics can still be expressed in the required form.

To prove this point, we decided to implement our best-performing method, the $M_{W(X,S,\widehat{Y})}$ for the Equality of Odds (EoO) metric : 
\[
\text{EoO}(f, \mathbb{P}, S)= |\mathbb{P}(\widehat{Y}=1|S=1 \land Y=1) - \mathbb{P}(\widehat{Y}=1|S=0 \land Y=1)|
\]

Note that similarly to the Disparate Impact, which is the multiplicative counterpart of the Disparate Parity, we could have taken the multiplicative definition of the EoO. However, we choose the additive definition because the multiplicative case is trivial for us, as we could have directly applied our DI-fairwashing method on the $Q_{n,Y=1}$. 

\begin{figure}[t!]
 \centering
 \includegraphics[width=0.35\textwidth, height=0.35\textwidth]{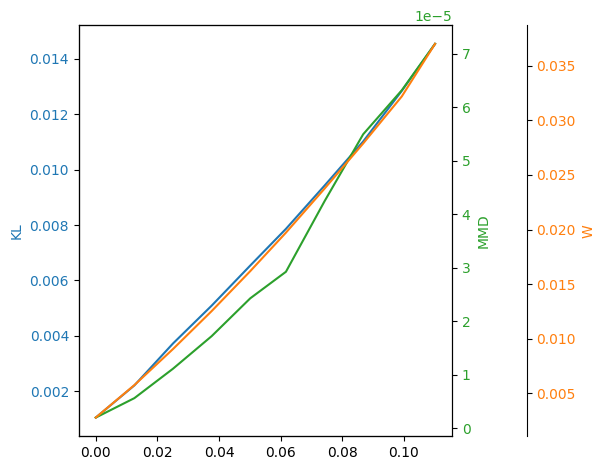}
 \caption{Distribution shift of Wasserstein distance, KL divergence and MMD, when constraining the Equality of Odds (EoO) fairness metric on the Adult dataset using the $M_{W(X,S,Y)}$ method.}
 \label{app:fig:eoo_line}
\end{figure}

The only difference to the matching method $M_{W(X,S,\widehat{Y})}$ going from DI constraint to EoO constraint is, following the notation of Section~\ref{sec:wass_matching}, iteratively from maximizing the left part of Eq.~\ref{app:eq:m_di_to_eoo} to its right part.

\begin{equation} \label{app:eq:m_di_to_eoo}
 \frac{\text{DI}(\tau_{i_0}(Z^j)) - \text{DI}((Z^j))}{\|\tau_{i_0}(Z^j) - Z^j\|} \to - \frac{\text{EoO}(\tau_{i_0}(Z^j)) - \text{EoO}((Z^j))}{\|\tau_{i_0}(Z^j) - Z^j\|}
\end{equation}

The minus sign comes from the difference between the fairness metric : an independence toward the sensitive variable $S$ for the Disparate Impact implies $DI=1$, we therefore try to maximize the DI. On the other hand, independence for the EoO implies $EoO=0$, leading us to minimize this criterion (i.e., maximizing minus the criteria). We illustrate this capacity in Fig.~\ref{app:fig:eoo_line}.

\section{Comparison with fairness manipulation via the Stealthily Biased Sampling (\texttt{SBS}) method}

\subsection{Explaination of the \texttt{SBS} method} 

The method designed by \cite{fukuchi2020faking} minimizes the distribution shift measured by $W(X)$ under a fairness constraint on the Disparate Parity (DP): 
\[
\text{DP}(f, \mathbb{P}, S) = |\mathbb{P}(\widehat{Y}=1|S=1) - \mathbb{P}(\widehat{Y}=1|S=0)|
\]

Notably, the method does not allow specifying a target threshold $t$ for the fairness criterion. Instead, the authors designed their sampling procedure to produce a perfectly fair dataset, such that $\text{DP} = 0$ in expectation. The only tunable hyperparameter is the common acceptance rate for positive outcomes, denoted by $\alpha:=\mathbb{P}(\widehat{Y}=1|S=1)=\mathbb{P}(\widehat{Y}=1|S=0)$.

This lack of flexibility in selecting a targeted DP complicates direct comparisons. As demonstrated in our paper, achieving fairness solely to pass compliance checks (i.e., fairwashing) often remains detectable by our statistical tests. To evaluate robustness, we progressively relax the fairness constraint until samples evade detection. Such adaptive calibration is not feasible with their approach.

One practical advantage of their method is that it outputs individual sampling probabilities rather than a fixed dataset, similar to our \texttt{Entropic} approach. This allows us to resample and generate distributions with varying degrees of fairness. Their reported results were obtained via grid search over $\alpha$ values, as illustrated in Fig.~\ref{app:fig:st_b_s}. Consequently, to benchmark their method, we either had to identify the optimal $\alpha$ minimizing distribution shift or evaluate performance across all tested $\alpha$ values.

This method’s high computational cost, already acknowledged by the authors in a subsequent paper \cite{yamamoto2024fast}, is a notable limitation. Due to these computational constraints, we applied their method exclusively on the Adult dataset, as experiments on larger datasets failed to complete within reasonable time frames.

\subsection{Result of the \texttt{SBS} method in our audit.} 

In this section, we evaluate on the Adult dataset: (1) the distribution shift incurred when creating a fair distribution ($\text{DI}(f, Q_t, S) > 0.8$) and (2) the maximum achievable DI without detection by our statistical tests. The first comparison (1) is well-aligned with the purpose of the method, which targets compliance. However, the second (2) inherently disadvantages their approach, as it was not designed to trade off fairness against detectability.

\begin{table}[tb!]
 \centering
 \caption{
 Distribution shift induced by different fairwashing methods on the Adult dataset at a target Disparate Impact threshold $\text{DI}(f, Q_t, S) \geq 0.8$. Wasserstein and KL divergence metrics are computed on the joint distribution $(X,S,\widehat{Y})$ as well as on $(S,\widehat{Y})$ only. Costs are evaluated on the projected dataset (or on the projected distribution for \texttt{Entropic} and \texttt{SBS} methods). Lower values indicate smaller distribution shifts.
 }
 \label{tab:STBS_comp_metric}
 
  \resizebox{\textwidth}{!}{
 \begin{tabular}{l ccc ccc ccc}
 \toprule
 & \multicolumn{9}{c}{Unbiasing Methods}     \\
  \cmidrule(r){2-10}
  
 Dataset & \texttt{SBS} & Grad\_p & Grad\_b & Grad\_p\_1D & Grad\_b\_1D & Rep $(S,\widehat{Y})$ & $M_{W(X,S,\widehat{Y})}$ & Entr\_b & Entr\_p \\
 \midrule
 $W(X,S,\widehat{Y})$ & 0.91 & 0.10 & 0.08 & 0.13 & 0.09 & \textbf{0.05} & \textbf{0.06} & 0.28 & 0.35 \\
 $W(S,\widehat{Y})$ & \textbf{0.00} & 0.09 & 0.08 & 0.09 & 0.08 & 0.05 & 0.05 & 0.08 & 0.09 \\
 $\text{KL}(X,S,\widehat{Y})$ & 0.73 & $\infty$ & $\infty$ & $\infty$ & $\infty$ & $\infty$ & 0.03 & \textbf{0.02} & 0.03 \\
 $\text{KL}(S,\widehat{Y})$ & 0.73 & 0.02 & 0.02 & 0.02 & 0.02 & 0.03 & 0.03 & 0.02 & 0.03 \\
 \bottomrule
\end{tabular}}
 
\end{table}


Regarding $d(Q_n, Q_t)$, their method, like our \texttt{Entropic} approaches, produces sampling probabilities rather than a direct sample. This yielded strong performance on $W(S,\widehat{Y})$, but it underperformed on $\KL(S,\widehat{Y})$ and did not stand out on $W(X,S,\widehat{Y})$. For $\KL(X,S,\widehat{Y})$, it was less competitive, though it notably avoided divergence to infinity, making it one of the more globally competitive Wasserstein-based methods.

\begin{table*}[tb!!]
 \centering
 \caption{Highest undetected achievable Disparate Impact for the Adult dataset, for each sample size (S Size) and fairwashing method. The symbol -- indicates that some methods couldn't reach a DI improvement. To emphasize the best method to use in order to deceive the auditor, we put the DI achieved in bold when one or two over-performed the others. We remind that the original DI of our Adult dataset is $0.30$.}
  \label{tab:highest_undetected_unbiasing_sbts}
 \resizebox{\textwidth}{!}{
 
 \begin{tabular}{ll c cccc cccc}
\toprule
Dataset & S size (\%) & \texttt{SBS} & Grad\_p & Grad\_b & Grad\_p 1D & Grad\_b 1D & Rep $(S, \widehat{Y})$ & Entr\_b & Entr\_p & $M_{W(X,S,\widehat{Y})}$ \\
\midrule
ADULT & 10 & 0.47 & 0.47 & 0.43 & 0.49 & 0.44 & 0.50 & \textbf{0.54} & 0.52 & \textbf{0.54} \\
  & 20 & -- & 0.39 & 0.40 & 0.38 & 0.39 & 0.41 & \textbf{0.42} & 0.41 & \textbf{0.42} \\
\end{tabular}%
}
\end{table*}

Conversely, as shown in Table~\ref{tab:highest_undetected_unbiasing_sbts}, the method is not suitable for maximizing fairness without detection. To assess this, we computed 500 samples from each tuple sample size, $\alpha$ and observed whether the samples passed our statistical tests. Across $500*10*2$ samples (10 $\alpha$ and 2 sample size), 5 samples passed the 7 statistical tests, they were all for $\alpha = 0.25$ and sample size of 10\% (instead of 20\%). 

\subsection{Technical insecurities}

The CMake version the authors used was 2.8 which is no longer supported with CMake (oldest version supported is 3.5) ; when changing in the CMake file the minimum version to 3.5, we had encountered another error with their (CMake\_policy(set CMP0048 OLD)) which is no longer supported as well, we change it to the new version. 

\begin{table}[tb!]
 \centering
 \caption{Old and new results of the \texttt{SBS} method on the Adult dataset, '--' means that the new version has \textbf{exactly} the same result as the old one.}
  \label{app:tab:st_b_s}
 \begin{tabular}{ll ccccc}
 \toprule
  & Version & Accuracy &	DP	& WD on Pr[x]&	WD on Pr[x|s=1]	& WD on Pr[x|s=0] \\
  \midrule
Baseline & Old &	0.851	& 0.1824	&22.1638	&25.6454	&35.0421 \\
   & New &	0.85115	&--	&--	&--	& -- \\
Case-control& Old	&NaN	& 0.0250	&23.9060	&22.5855	&37.9543 \\
& New & NaN	& 0.0243	& 23.2002	& 23.3179	&37.8548\\
Stealth	& Old & NaN	& 0.0712	& 23.6396	&24.2404	&36.1657 \\
&New	&NaN	& 0.0708	& 24.1415	&25.1640	&35.5028\\
 \bottomrule
 \end{tabular}
\end{table}

\begin{figure}[tb!]
  \centering
  \begin{subfigure}[t]{\textwidth}
    \centering
    \includegraphics[height=1.3in]{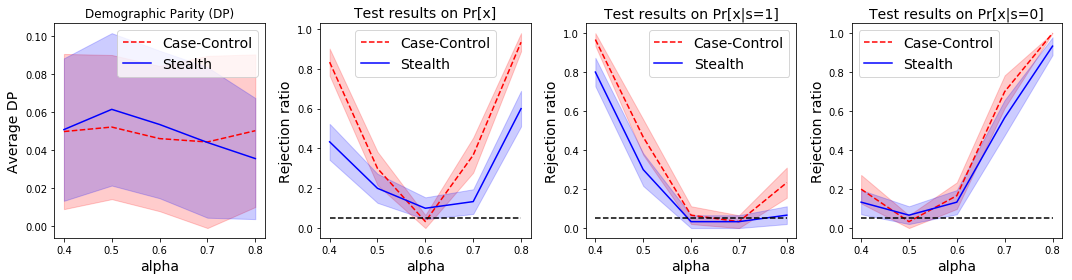}
    \caption{Original version}
    \label{fig:STBS_old}
  \end{subfigure}
 \\
  \begin{subfigure}[t]{\textwidth}
    \centering
    \includegraphics[height=1.3in]{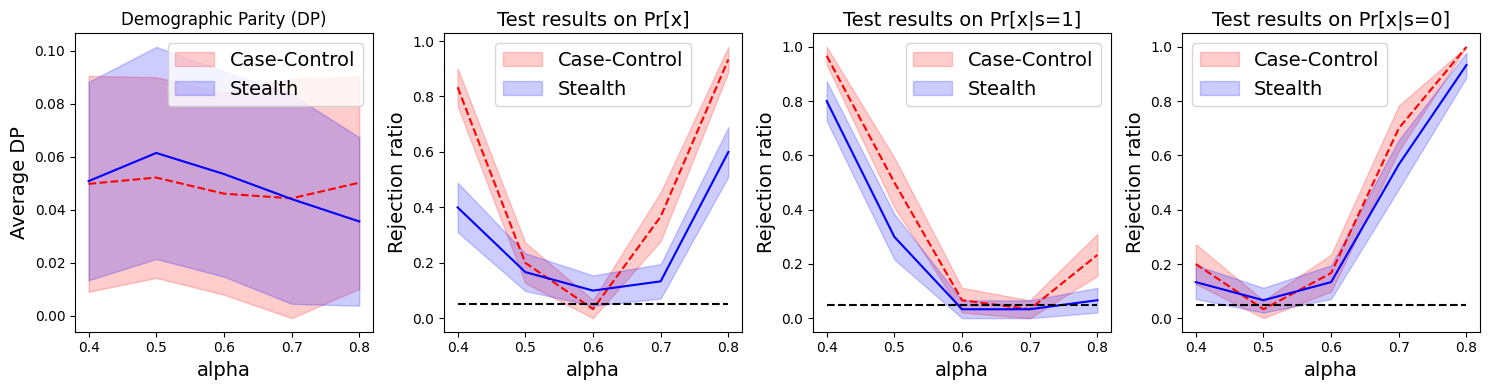}
    \caption{Latest version}
    \label{fig:STBS_NEW}
  \end{subfigure}
  \caption{Original (Old) and latest (New) results for their synthetic datasets, obtained over 30 runs for several values of $\alpha$, confirm a negligible impact on performance, with no observable difference between the two versions.}
  \label{app:fig:st_b_s}
\end{figure}

According Table~\ref{app:tab:st_b_s}, we observed a slight mismatch between the value they obtained and the value we obtained running exactly their code due to a newer version of libraries. Then we consider that our result might actually be more representative. However, Fig.~\ref{app:fig:st_b_s} indicates that the aggregated results are very similar. We then provide a comparison using our implementation rather than their fine-tuned Sliced Wasserstein Distance method \cite{yamamoto2024fast}, with a negligible impact on performance and a significant computational speed-up.

\section{Extension to other data type}\label{app:sec:extension}
The method we develop is originally meant to handle tabular data. However we propose some natural direction to extend this work to text or images. The distances used to evaluate
Wasserstein distance or the Maximum Mean Discrepancy (MMD) relies on the inherent informative information between individual within the input space, in another word they rely on the fact that the distance between individual is proportional to their semantic similarity. This hypothesis is practically never rejected on tabular data (with the $L_2$ distance, for instance), but it might be on images or token distributions. 
We will first evaluate our method based on $W(X)$ or $\text{MMD}(X)$ to detect fraud attempt and expect the method to achieve a lower efficiency because $d(Q_t, Q_n)$ is hardly related to the semantic meaning of the images. Thus a fairwashing manipulation might not change this distance distribution.

Hence we embed the images in another space, where the regular distances have semantic meanings. The construction of such a space has already seen numerous works, including Principal Component Analysis, using the latent space of Auto-encoder or Variational Auto-Encoder, or using the latent space of Convolutional Neural Network classifiers. Using such space, which we call descriptor $D$, have become common practice after the introduction of the Fréchet Inception Distance (FID)\cite{heusel2018ganstrainedtimescaleupdate}. We define the function $E$ such as 
\begin{align*}
E\colon\; & \mathbb{R}^N \to \mathbb{R}^m {\hspace{2cm}N,m\in\mathbb{N}, N\gg m } \\ & X \mapsto E(X) = D
\end{align*}

and set $E(Q):=\{E(X)|X\in Q\}$ if $Q$ is a distribution.
We choose in the following latent features given by the CNN classifier.
\subsection{Experimental settings}

We audited the CelebA dataset \cite{liu2015faceattributes}: predicting the attractiveness, with the sensitive attribute being having heavy makeup. We note here that we choose this sensitive attribute instead of others for mainly two reasons: 
\begin{enumerate}
    \item Low DI : 0.4 on the whole dataset
    \item Representativeness : Similarly to what we saw for the BAF dataset having a low probability of $\mathbb{P}(Y=1)=0.01$, If the sensitive variable was too rare, then detecting a modification on $X$ would be impossible (for tabular or non-tabular data) 
\end{enumerate}
Note also that the variable \textit{young} would have been another viable candidate. 

The fairwashing method used in those experiments is the Wasserstein-based matching method $M_{W(X,S,Y)}$. We fine-tune 3 CNN models: an InceptionV3 \cite{szegedy2015rethinkinginceptionarchitecturecomputer}, a ResNet18 and a ResNet101 \cite{he2015deepresiduallearningimage} on CelebA and select part of the test set to audit, on this subset we observe respectively a DI of $0.34$, $0.35$ and $0.35$. The malicious auditee, aware that the statistical tests on the covariates $X$ might not be on the pixels of the images, but on the descriptor of the images, could minimize $d(E(Q_n), E(Q_t))$ instead of $d(Q_n, Q_t)$. Therefore, we consider 6 different fairwashing scenarios given (1) the network choice amongst ResNet18, InceptionV3 and ResNet101 which implies different descriptors' space and (2) if the auditee optimized the Wasserstein-based matching method on the pixel's space or on the latent space of those models. We first investigate the use of statistical tests directly of the pixels' space. Secondly, for each of the above scenario, we use statistical tests based of the latent space of the CNN. We remind here that (1) in term of complexity, the CCN are ranked as follow: ResNet18 ($11$ million parameters) $<$ InceptionV3 ($27$ million parameters) $<$ ResNet101 ($44$ million parameters), (2) the latent space of the CNN is the space at the hidden layer before the last linear layer, for the three models above's latent space share the same dimension size of 1000. 

\subsection{Topics of Interest and Answers}

\begin{table}[t!]
  \centering
  \scalebox{0.9}{\begin{tabular}{ll lll lll}
    \toprule
    & & \multicolumn{6}{c}{Fairwashing minimization objective} \\     
     Descriptors  & Size (\%) & 18 & 101 & v3 & 18 pixels & 101 pixels & v3 pixels \\ 
     \midrule
     ResNet18 & 10 & $\geq0.95$ & $\geq0.95$ & $\geq0.95$ & $\geq0.95$ & $\geq0.95$ & $\geq0.95$\\
      & 20 & $[0.6-0.7[$ & $[0.5-0.55[$ & $[0.5-0.55[$ & $[0.4-0.5[$ & $[0.4-0.5[$ & $[0.4-0.5[$ \\

     Inceptionv3 & 10 & $\geq0.95$ & $\geq0.95$ & $[0.8-0.95 [$ & $\geq0.95$ & $[0.8-0.95 [$ & $[0.8-0.95 [$\\
     & 20 & $[0.55-0.6 [$ & $[0.5-0.55 [$ & $[0.5-0.55 [$ & $[0.4-0.5 [$ & $[0.4-0.5 [$ & $[0.4-0.5 [$ \\

     ResNet101 & 10 & $\geq0.95$ & $\geq0.95$ & $\geq0.95$ & $\geq0.95$ & $\geq0.95$ & $[0.7-0.8 [$\\
     & 20 & $[0.55-0.6 [$ & $[0.5-0.55 [$ & $[0.55-0.6 [$ & $[0.4-0.5 [$ & $[0.4-0.5 [$ & $[0.4-0.5 [$ \\
    \bottomrule
  \end{tabular}}
    \caption{Highest DI without being detected for the CelebA Dataset using the matching fairwashing method based on different minimization objective testing on the descriptors which are the latent space of the different models, for sample size of 10\% and 20\%. The different scenarios are the following: 18, 101 and v3 are respectively a ResNet18, a ResNet101 and an Inceptionv3 optimized on their latent space ; the 18 pixels, 101 pixels and v3 pixels are the methods optimized on the pixel space (even if they have the same objective, they are different because the prediction of each network might be different).}
      \label{app:tab:celebA_descri}
\end{table}

In this subsection, we present key questions of interest and provide direct answers. These answers are supported by selected (cherry-picked) results for illustrative purposes, we refer to Table~\ref{app:tab:celebA_descri} for complete results on the descriptors analysis results. We address the following questions:

\begin{enumerate}
    \item Are the statistical tests presented in our paper for tabular data relevant for non-tabular data?
    \begin{itemize}
        \item Yes. We consider several data-agnostic statistical tests, such as $W(S,Y)$, $\text{MMD}(S,Y)$, KL divergence, and the KS test. These tests remain applicable regardless of data type.
        \item What about tests based on $W(X,S,Y)$ or $\text{MMD}(X,S,Y)$, which rely on $W(X)$ or $\text{MMD}(X)$ where $X$ is high-dimensional?
        Tests conducted directly in pixel space can be useful. For example, fairwashing attempts with a sampling size of 20\% were detected using tests applied directly to raw pixel values. In contrast, with a sampling size of 10\%, and fairwashing to a Disparate Impact (DI) of 0.95, the null hypothesis $\mathcal{H}_0$ was not rejected, this indicates that these manipulations may go undetected at smaller sampling sizes.
    \end{itemize}

    \item Are statistical tests based on learned descriptors more effective?
    \begin{itemize}
        \item Yes. Descriptor-based tests can detect fairwashing even with smaller samples. For instance, fraud attempts were identified using only 10\% of the CelebA dataset (noting that detection becomes harder with smaller samples).
        \item However, the auditee could potentially optimize their manipulation based on the descriptor used by the auditor, rendering these tests ineffective again at the 10\% sample size.
    \end{itemize}

    \item Is the choice of the descriptor impactful ?
    \begin{itemize}
        \item Yes, the results do depends on both the auditee choice of descriptor as well as the descriptor used in the statistical tests.
        \item Yes for the manipulation: it was harder to detect manipulation based on the ResNet18 descriptors, and the easiest was the manipulation based on the ResNet101 descriptors. For instance, for the sampling size of 20\%, even while testing using the ResNet101 descriptors, for a fairwashing at $DI=0.55$, it was undetected when optimizing using ResNet18 descriptors but detected when optimizing using ResNet101's one (by ``detected'', we mean that across multiple samples, 50 in this case, the  hypothesis $\mathcal{H}_0$, i.e., the hypothesis that the sample and original distributions are the same, was rejected every time.). 
        \item Yes for the fraud detection: Statistical tests based on the ResNet18 was more easily fooled by manipulation. To support this claim, we refer to for example to the last three columns of the Table~\ref{app:tab:celebA_descri} where with optimization on the image pixels, for a 10\% sampling size, no fairwashing method was detected even with $DI=0.95$.
    \end{itemize}

    \item Is there a difference using statistical tests based on the latent space the auditee's fairwashing method optimized on?
    \begin{itemize}
        \item No, our results are not conclusive enough to answer this question positively. For InceptionV3 and ResNet101, we did not observe a significant difference.
        \item That being said, in our experiments, with a 20\% sampling size and fairwashing to $DI=0.60$, only the auditee optimizing on the same descriptors (ResNet18) was able to generate an undetected sample when the test was based on those same descriptors.

        \item Importantly, in practice, it is unlikely that the supervisory authority would use the same descriptors as the auditee. Even if the authority had full access to the auditee's network (which is rare, since this would go beyond API access), they may deliberately avoid using the same descriptors to prevent optimization-based circumvention.
    \end{itemize}
\end{enumerate}

In conclusion, on non‑tabular modalities, running statistical tests directly on raw signals (in our cases pixels) is not useless, but tests in a learned descriptor space are markedly more sensitive. The choice of descriptor is critical: tests based on higher‑capacity, semantically rich encoders (e.g., ResNet101) are substantially more robust to manipulations. We therefore recommend that supervisory authorities apply statistical tests both on the raw data and in a high‑quality descriptor space. For text datasets, though not evaluated here, a natural first descriptor we would recommend is the CLS embedding from a BERT‑style model~\cite{devlin2019bertpretrainingdeepbidirectional}, we leave this for a further work.

\section{Discussion concerning accuracy}
\label{app:sec:accuracy}

While having the best prediction accuracy was not the goal of experiments, we still achieve reasonable accuracy learning with the {\it ScheduleFree} optimizer \cite{defazio2024road}. We defined the logit threshold based ground truth mean : $ l_{th} := \min_{ l \in ]0,1[} | \mathbb{E}(\widehat{Y_l}) - \mathbb{E}(Y)|$ with $\widehat{Y_l} = \{ f(x) > l | x\in \mathcal{D}\}$. This was especially necessary for the BAF dataset, where the learning task is basically an anomaly detection task, and $\mathbb{E}(Y) \approx 0.01$. 

In our work, the predictor accuracy is not the primary object of study. It serves as a representative biased AI system subject to auditing. Similarly, we deliberately did not apply bias-mitigation strategies, as a fairness-compliant system would have no incentive to engage in fairwashing.

For completeness, we now report predictive accuracies (per dataset):
\begin{itemize}
    \item Adult Census Income: 84\%
    \item Folktables Employment: 77\%
    \item Folktables Income: 88\%
    \item Folktables Mobility: 84\%
    \item Folktables Public Coverage: 73\%
    \item Folktables Travel Time: 72\%
    \item Bank Account Fraud: 98\% (high accuracy due to strong class imbalance)
\end{itemize}

While predictive performance could likely be improved (we used similar architectures across datasets), maximizing accuracy was not the objective of this work.

\section{Model sensibility}

Except for the \texttt{Grad} methods, which require differentiable models, our fairwashing procedures are largely model-agnostic. For instance, classical tabular methods (e.g., gradient-boosted trees) could have been used for tabular datasets without fundamentally changing the conclusions. Across random seeds, accuracy typically varies within $\pm 2\%$. However, Disparate Impact (DI) can vary more substantially (sometimes by more than $0.1$), especially for highly biased models. This variability motivated our use of multiple datasets.

\section{Extension to other models than binary classifier fairwashing}

Using the Wasserstein gradient-based approach, we modify the model output $f(x) \in \{0,1\}$. In practice, to better monitor convergence, we operate on the neural network’s logits rather than on the hard labels. After applying a sigmoid function, the output is interpreted as a probability $f(x) \in [0,1]$, which provides a smoother signal for optimization. This allows the constraints to be imposed directly on the logits 
(or equivalently on the corresponding probabilities). Importantly, this shows that the same methodology naturally extends beyond binary classification to regression and other continuous-output settings.

More generally, both \texttt{Entropic} methods and \texttt{Wasserstein gradient} methods are designed to operate directly on continuous outcomes $Y$. The \texttt{Replace} and \texttt{Matching} methods can also be adapted to continuous $Y$, although at a significantly higher computational cost.

Overall, the fairwashing methods proposed in this paper can therefore be used to simulate malicious manipulation by an auditee in regression settings, or in scenarios where fairness metrics are applied to logits rather than to discrete outcomes. For example, a regulator might choose to evaluate fairness at the logit level using a criterion such as Equality of Odds (see Sec.~\ref{app:sec:other_fairness_metric}). Since neural networks are not inherently calibrated, this raises the question of whether such an evaluation framework would be legally admissible.

\section{Auxiliary results}

\begin{proposition}\label{prop:extproje}
    Consider the following minimization problem 
\begin{align}\label{P1}
    \min W_2^2(P,Q_n) \text{ such that } \int_E \Phi(x)dP(x) \geq t.
\end{align}
Then $Q_t$ is optimal for \eqref{P1} if, and only if, it is defined as the push-forward
\[ Q_t = {T_{\lambda^\star}}_{\#}Q_n \]
 where 
$ T_\lambda(y) \in {\rm arg}\min_{x}  \left\{\|x-y\|^2-\lambda^T \Phi(x) \right\}   $ and 
and then $\lambda^\star\in\R^k_{\geq0}$ solves 
\begin{itemize}
    \item $\int_E \Phi(T_{\lambda^\star}(x))dQ(x) \geq t $,
    \item and $\langle \lambda^\star ,  t-t_{\lambda^\star} \rangle=0$ 
\end{itemize} 
\end{proposition}

\section{Proofs}

\subsection{Proof of Proposition~\ref{th:FWKL}}
\begin{proof}
    
Theorem~\ref{thm:discrete:reweigth:multidim} implies the existence of a distribution $Q_t$ such that \[DI(f,Q_t,S)= \frac{\lambda_0 + \delta_0}{\lambda_1 - \delta_1}\frac{n_1}{n_0}=t_1. \]
 We have
\[\Delta_{DI} =
 \frac{n_1}{n_0}\bigg( \frac{\lambda_1 \delta_0 + \lambda_0 \delta_1)}{(\lambda_1 - \delta_1)\lambda_1}\bigg)
\] 
Among all possible solutions, we privilege the two solutions described in the Proposition. Knowing the new DI desired, we can obtain a set of solution for $\delta_0$ and $\delta_1$.
 \end{proof}

\subsection{Proof of Theorem \ref{thm:empirical}}

\begin{proof}
    First, notice that the definition of $T_\lambda$ implies
\begin{align*}
    W_2^2(Q_n,{T_\lambda}_\#Q_n)&\leq\int_E\|T_\lambda(y)-y\|^2dQ_n(y)\\
    &=\int_E\|T_\lambda(y)-y\|^2dQ_n(y) + \frac{1}{n}\left(\sum_{i=1}^n\lambda^\top\Phi(T_\lambda(Z_i))-\sum_{i=1}^n\lambda^\top\Phi(T_\lambda(Z_i))\right)\\
    &=\int_E\|T_\lambda(y)-y\|^2-\lambda^\top\Phi(T_\lambda(y))dQ_n(y) + \int_E\lambda^\top\Phi(y)d{T_\lambda}_\#Q_n(y)\\
    &=\int_E\inf_x\left\{\|x-y\|^2-\lambda^\top\Phi(x)\right\}dQ_n(y) + \int_E\lambda^\top\Phi(y)d{T_\lambda}_\#Q_n(y)\\
    &=\int_E(\lambda^\top\Phi)^c(y) dQ_n(y) + \int_E\lambda^\top\Phi(y)d{T_\lambda}_\#Q_n(y).
\end{align*}
Strong duality of the Kantorovich problem, see \cite{sant}, guarantees that this inequality is indeed an equality. Since our equality constraint is linear, a necessary and sufficient condition for $P^*$ to be a minimizer, see \cite{Peypouquet}, is finding Lagrange multipliers $\lambda_1,\ldots,\lambda_k\in\R$ such that 
    \begin{align*}
        &\sum_{i=1}^k\lambda_i\nabla g_i(P^*)\in \partial f(P^*)\quad\text{ (extremality condition)}\\ 
        & g(P^*) = 0 \quad\text{ (feasibility)}
    \end{align*}
    where $g(P) = \int_E \Phi(x)dP(x) - t$ and $f(P) = W_2^2(P,Q_n)$.
    The subgradient of $f$ is given, see Proposition 7.17 in \cite{sant}, by the set of Kantorovich potentials between $P^*$ and $Q$:
    \begin{align}
        \partial f(P^*) = \left\{ \phi\in C(E)\mid \int\phi dP^*+ \int\phi^c dQ = W_2^2(P^*,Q)\right\}.
    \end{align}
Our computations above prove the extremality condition for $P^*={T_{\lambda^\star}}_\#Q_n=\frac{1}{n} \sum_{i=1}^n \delta_{T_{\lambda^\star}(Z_i)}$ since $\nabla g_i(P) = \int\Phi dP$. The feasibility condition for the empirical measure $Q_n$ is to find $\lambda^\star$ such that
\begin{align}
    t = \int_E\Phi(y)d{T_{\lambda^\star}}_\#Q_n(y)=\frac{1}{n}\sum_{i=1}^n\Phi(T_{\lambda^\star}(Z_i)).
\end{align}

\end{proof}

\subsection{Proof of Proposition \ref{prop:extproje}}

\begin{proof}
Let $g$ be the continuous function $g(P) = t - \int_E \Phi(x)dP(x)$ and $f(P) = W_2^2(P,Q)$. The set $\{P\in\mathcal{M}(E)\mid \int_E \Phi(x)dP(x) \geq t\} = g^{-1}([0,\infty))$ is closed for the weak convergence as $[0,\infty)$ is closed. Then the projection problem is well-definedd. 
Before applying the Lagrange multiplier theorem, we must verify Slater's condition. By continuity of $\Phi_i$ and compacity of $E$ we can consider, for $i = 1, \ldots, k$, $x_0^i\in E$ such that $\Phi_i(x_0^i)=\min_{x\in E}\Phi_i(x)$. Take $\alpha\in\R$ such that $\max_{1\leq i \leq k}t_i/\Phi_i(x_0^i)<\alpha.$ Then $\Bar{P} = \alpha\delta_{x_0^i}$ satisfies $g_i(\Bar{P})<0$ for $i = 1, \ldots, k$. The Lagrange multipliers theorem guarantees that $P^*$ is optimal for \ref{P1} if, and only if, there exists $\lambda_1,\ldots,\lambda_k\geq0$ such that 
    \begin{align*}
        &\sum_{i=1}^k\lambda_i\nabla g_i(P^*)\in \partial f(P^*)\quad\text{ (extremality condition)}\\ 
        & g(P^*) \leq 0 \quad \text{ and } \lambda_ig_i(P^*)=0 \text{ for all } i = 1, 2, \ldots, k \text{ (feasibility)}
    \end{align*}

The proof of the extremality condition is completely analogous to the proof of Theorem \ref{thm:empirical}, replacing $Q_n$ by $Q$.
To conclude, we need to find $\lambda^\star\in\R^k_{\geq0}$ such that the feasibility condition is satisfied:
\begin{align}
    t \leq\int_E \Phi(T_{\lambda^\star}(x))dQ(x) \text{ and } {\lambda^\star}^\top\left(t-\int_E\Phi(T_{\lambda^\star}(x))dQ(x)\right) = 0.
\end{align}
\end{proof}

\subsection{Joint convexity of the Wasserstein distance under mixture-preserving coupling}\label{app:sec:proof_ineq}

Let \( Q_n \) and \( Q_t \) be probability distributions over \( \mathcal{X} \times \{0,1\} \), where \( X \in \mathcal{X} \) denotes the data and \( S \in \{0,1\} \) is a binary group attribute.

For each \( s \in \{0,1\} \), define the conditional distributions:
\[
Q_{n,s} := Q_n(\cdot \mid S = s), \quad Q_{t,s} := Q_t(\cdot \mid S = s),
\]
and let \( \pi := Q_n(S = 1) \in [0,1] \). Then, define the marginal (mixture) distributions over \( \mathcal{X} \) as:
\[
\mu := \pi Q_{n,1} + (1 - \pi) Q_{n,0}, \quad 
\nu := \pi Q_{t,1} + (1 - \pi) Q_{t,0}.
\]

We prove the inequality:
\[
W_2^2(\mu, \nu) \leq \pi W_2^2(Q_{n,1}, Q_{t,1}) + (1 - \pi) W_2^2(Q_{n,0}, Q_{t,0}).
\]

\begin{proof}

Let \( \gamma_1 \in \Pi(Q_{n,1}, Q_{t,1}) \) and \( \gamma_0 \in \Pi(Q_{n,0}, Q_{t,0}) \) be couplings between the corresponding conditionals. Define the coupling:
\[
\gamma := \pi \gamma_1 + (1 - \pi) \gamma_0.
\]

Then \( \gamma \in \mathcal{P}(\mathcal{X} \times \mathcal{X}) \), and its marginals are:
\[
\gamma^X = \pi Q_{n,1} + (1 - \pi) Q_{n,0} = \mu, \quad 
\gamma^Y = \pi Q_{t,1} + (1 - \pi) Q_{t,0} = \nu.
\]
Thus, \( \gamma \in \Pi(\mu, \nu) \) is a valid coupling between \( \mu \) and \( \nu \).

Now compute the transport cost under $\gamma$:
\[
\int_{\mathcal{X} \times \mathcal{X}} d(x,y)^2 \, d\gamma(x,y)
= \pi \int d(x,y)^2 \, d\gamma_1(x,y)
+ (1 - \pi) \int d(x,y)^2 \, d\gamma_0(x,y),
\]
(Because the distance is an integrable function, we can use the linearity of the Lebesgue integral with respect to measures)
\[
= \pi W_2^2(Q_{n,1}, Q_{t,1}) + (1 - \pi) W_2^2(Q_{n,0}, Q_{t,0}).
\]

Since $W_2^2(\mu, \nu)$ is the infimum of such costs over all couplings in $\Pi(\mu, \nu)$, we obtain:
\[
W_2^2(\mu, \nu)
\leq \pi W_2^2(Q_{n,1}, Q_{t,1}) + (1 - \pi) W_2^2(Q_{n,0}, Q_{t,0}).
\]
\end{proof}

\section{Results with Simulated dataset}
\label{app:sec:simu}

\begin{figure}[ht!]
    \centering
    \begin{minipage}{0.49\textwidth}
        \centering
        \includegraphics[width=1\textwidth, height=4cm]{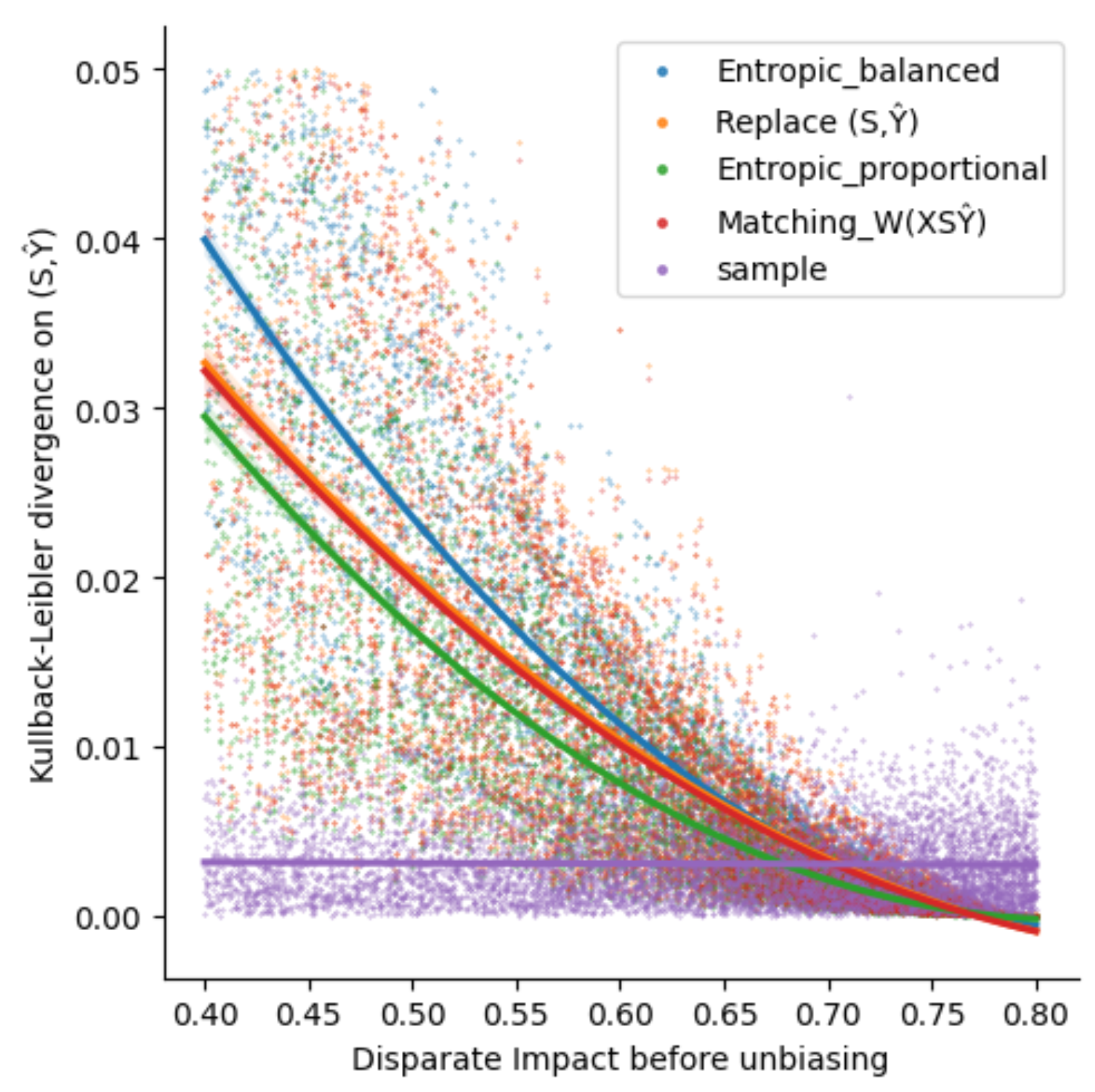}
    \end{minipage}
    \hfill
    \begin{minipage}{0.49\textwidth}
        \includegraphics[width=1\textwidth, height=4cm]{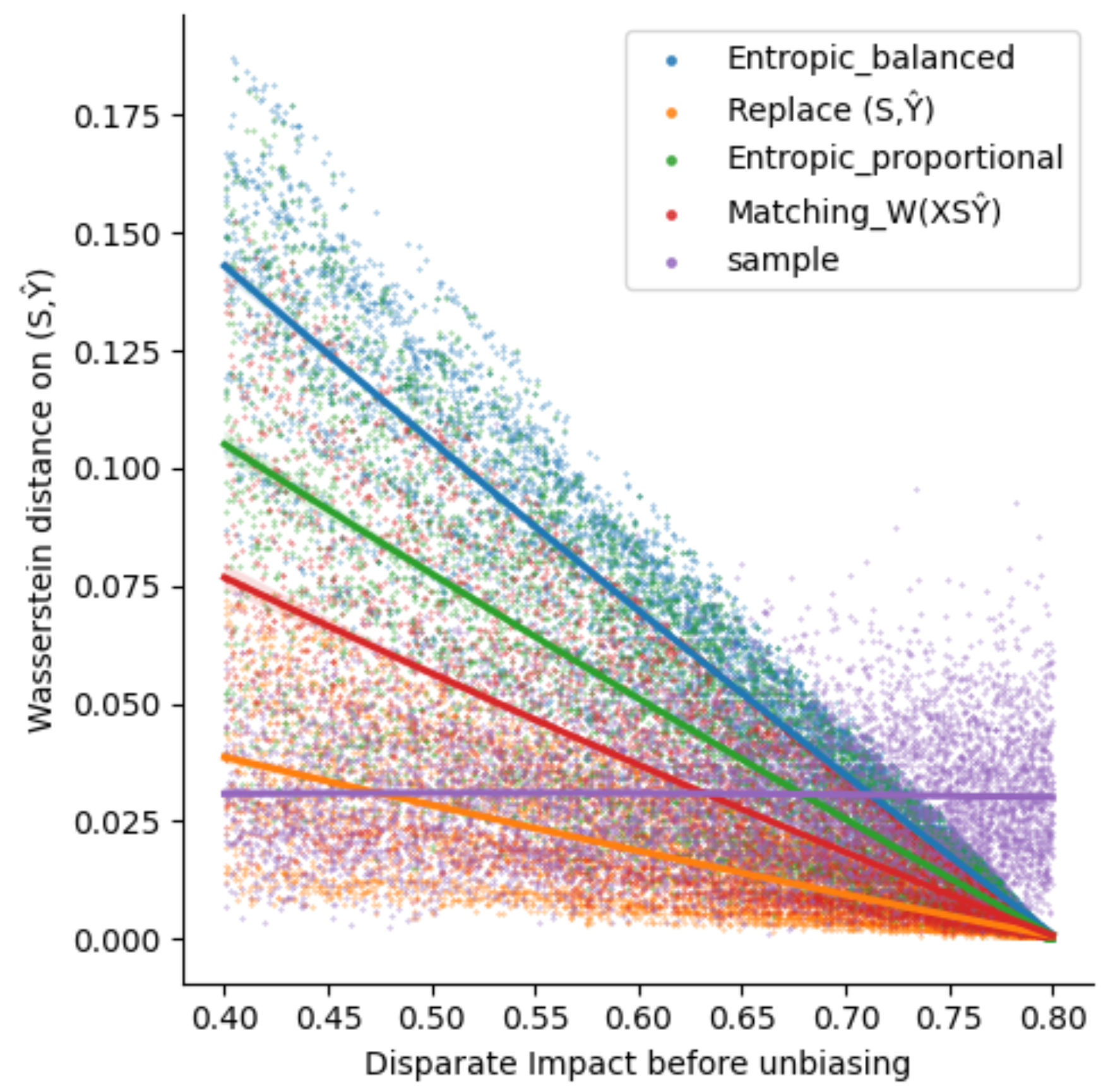}
    \end{minipage}
\caption{Logistic regression plots showing how the distance
(left: $\KL{(S,\widehat{Y})}$ and right:
$W_{(S,\widehat{Y})}$) between the original and 20\% of manipulated
datasets varies with the initial Disparate Impact for each fairwashing
method, with the manipulated dataset having DI = 0.8.
The sample results in the legend represent values from random samples
drawn from the original distribution $Q_n$.}
    \label{fig:simu_d}
\end{figure}

We create a simulated dataset to cover all possible cases where $ S \in \{0,1\}$ and $\widehat{Y} \in \{0,1\}$. The simulation parameters control $ \mathbb{E}(S)$, $ \mathbb{E}(\widehat{Y} \mid S = 0)$, and $\mathbb{E}(\widehat{Y} \mid S = 1)$, allowing us to represent a wide range of scenarios.
Fig~\ref{fig:simu_d} presents two logistic regression graphs illustrating how the distance between the complete original and 20\% of manipulated data evolves from the initial Disparate Impact (before debiasing), to reach a DI=0.8, for our different correction methods.
Methods with the highest KL and Wasserstein distance implies a high risk of being detected by a statistical test on the distribution. The lower the initial DI, the greater the change required to reach an acceptable DI (making fraud detection more likely).
When the original DI isabove $0.55$, the methods \texttt{Entropic}, \texttt{Replace} and \texttt{Matching} are equivalent in terms of KL divergence. Regarding the Wasserstein distance, they become equivalent for original DI values above $0.65$. Since the \texttt{Sample} method does not modify the original data, it preserves the distributional distances (KL and Wasserstein), and can be used as a reference: when the logistic regression score of a method is lower than that of \texttt{Sample}, we can infer that the modified dataset would not be detected as significantly different from the original according to these criteria. Among all methods, $M_{W(X,S,\widehat{Y})}$ with an original DI$\in[0.45,0.70]$ achieves the best trade-off between KL divergence and Wasserstein distance, reaching the required DI while keeping the modified distribution close to the original.

\section{Further studies on the impact of the sample size}
\label{sec:futher_studies}
In our conclusion, we recommended strongly to the referring authorities, that in order to prevent undetectable fraud, with appropriate statistical tests, requiring a bigger sample size is one of the single most important point. To further support this claim, we provide in this section a study on the sample size impact on the Adult dataset.

\begin{figure}[ht!]
    \centering  
    \includegraphics[width=0.5\linewidth]{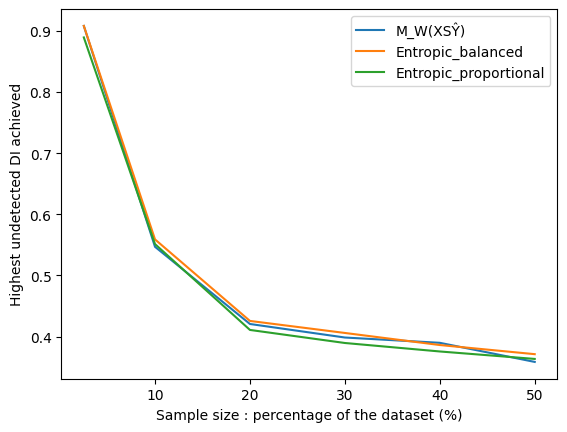}
    \caption{Highest Disparate Impact (DI) achieved without being detected by any statistical tests, constructed by different fairwashing methods, depending on the sample size, in the Adult dataset.}
    \label{app:fig:sample_size}
\end{figure}

Using the best performing fairwashing method ($M_{W(X,S,Y)}$, \\ \texttt{Entropic\_balanced} and \texttt{Entropic\_proportional}), we observe on Fig.~\ref{app:fig:sample_size} the highest DI achievable without being detected by our 7 statistical tests depending on the sample size required. 

\section{More information on the methods}

\subsection{Introduction}

In the methods section, we use a slightly different formulation of Disparate Impact than the standard legal definition. Specifically, we consider the \emph{unbounded} Disparate Impact ratio
\[
DI_{2}(f,\mathbb{P}, S) := \frac{\mathbb{P}(\widehat{Y} = 1 \mid S = 0)}{\mathbb{P}(\widehat{Y} = 1 \mid S = 1)},
\]
with the convention that
\[
\mathbb{P}(\widehat{Y} = 1 \mid S = 0) \leq \mathbb{P}(\widehat{Y} = 1 \mid S = 1),
\]
which is ensured by defining $S=0$ as the disadvantaged group. Under this convention, the bounded Disparate Impact used in the main text satisfies
\[
DI = \min\left\{ DI_2,\frac{1}{DI_2}\right\},
\]
as defined in \eqref{DI}.

With the \texttt{Entropic} method, it is possible to construct a distribution that achieves an exact target Disparate Impact, from which we can then sample. By contrast, the other methods operate on empirical datasets and are therefore constrained by the finite number of individuals $n = n_0 + n_1$. For example, with a dataset containing only $n=20$ individuals, it is generally impossible to achieve a threshold such as $0.9999 < DI_{\text{threshold}} < 1$.

As discussed in Remark~\ref{remark_ineq}, we therefore replace the equality constraint on $DI$ with an inequality constraint. Since our approach minimizes the Wasserstein distance, the optimization modifies only the minimal number of individuals required to exceed the desired threshold. In practice, the algorithm stops as soon as the constraint is satisfied, meaning that the threshold is crossed through a single modification of the dataset.

A natural question is whether the resulting distribution always satisfies $DI \geq DI_{\text{threshold}}$. From the construction, we know that
\[
DI_2 \geq DI_{\text{threshold}}.
\]
However, since the final metric is defined as $DI = \min\{DI_2,1/DI_2\}$, we must also ensure that
\begin{equation}
\dfrac{1}{DI_2} \leq DI_{\text{threshold}}
\quad \Longleftrightarrow \quad
DI_2 \leq \dfrac{1}{DI_{\text{threshold}}}.
\end{equation}

In practice, this condition is always satisfied when the dataset is sufficiently large.

To see this, recall that
\[
DI_2 = \frac{\lambda_0 n_1}{\lambda_1 n_0}.
\]
One way to increase $DI_2$ is to change the prediction of an individual with $(S,\widehat{Y})=(0,0)$ to $\widehat{Y}=1$. The resulting variation in $DI_2$ is
\begin{equation}
\label{app:eq:change_1}
\Delta_{DI_2} =
\frac{(\lambda_0 +1)n_1}{\lambda_1 n_0}
-
\frac{\lambda_0 n_1}{\lambda_1 n_0}
=
\frac{n_1}{\lambda_1 n_0}.
\end{equation}

Since $\lambda_1 = p_1 n_1$ with $p_1 \in (0,1]$ (and $p_1 > p_0$, implying $p_1 \neq 0$), we obtain
\[
\Delta_{DI_2} = \frac{1}{p_1 n_0}.
\]
As long as $p_1$ is not extremely small, the variation $\Delta_{DI_2}$ remains small.

In the rare case where $\mathbb{E}(\widehat{Y}) < 0.01$ and the dataset is very small, the concern may become relevant. A simple remedy is to artificially increase the dataset size while preserving the empirical distribution. Since $\Delta_{DI_2}$ decreases as the dataset grows (because $n_0$ increases while $p_1$ remains constant), we can apply deterministic oversampling.

For instance, duplicating each observation ten times while assigning each copy one tenth of the original probability preserves the empirical distribution while effectively multiplying the dataset size by ten. Let $\Delta_{{DI_2}_{\text{new}}}$ denote the corresponding change in $DI_2$. Then
\begin{equation}
\Delta_{{DI_2}_{\text{new}}}
=
\frac{(10\lambda_0 +1)10n_1}{100\lambda_1 n_0}
-
\frac{100\lambda_0 n_1}{100\lambda_1 n_0}
=
\frac{n_1}{10\lambda_1 n_0}
=
\frac{\Delta_{DI_2}}{10}.
\end{equation}

For completeness, we analyze the other possible modifications.

\begin{enumerate}

\item Changing the prediction of an individual with $(S,\widehat{Y})=(1,1)$ to $\widehat{Y}=0$ yields
\begin{equation}
\label{app:eq:change_2}
\Delta_{DI_2}
=
\frac{\lambda_0 n_1}{(\lambda_1 - 1)n_0}
-
\frac{\lambda_0 n_1}{\lambda_1 n_0}
=
\frac{\lambda_0 n_1}{n_0 \lambda_1 (\lambda_1 - 1)}.
\end{equation}

Again, oversampling reduces this variation:
\begin{equation}
    \Delta_{{DI_2}_{\text{new}}}
=
\frac{\lambda_0 n_1}{n_0 \lambda_1 (10\lambda_1 - 1)}
\approx
\frac{\lambda_0 n_1}{10 n_0 \lambda_1 (\lambda_1 - 1)}
=
\frac{\Delta_{DI_2}}{10}.
\end{equation}

The gradient-based Wasserstein methods only modify predictions, corresponding to \eqref{app:eq:change_1} and \eqref{app:eq:change_2}. However, the \texttt{replace} and \texttt{matching} methods may also alter the sensitive attribute $S$, which introduces two additional cases.

\item Changing an individual from $(S,\widehat{Y})=(1,0)$ to $(S,\widehat{Y})=(0,1)$ gives

\begin{equation}
    \label{app:eq:change_3}
    \Delta_{DI_2} = \frac{(\lambda_0+1)(n_1-1)}{\lambda_1(n_0+1)} - \frac{\lambda_0   n_1}{\lambda_1 n_0} = \frac{n_1n_0 - \lambda_0(n_1 + n_2) - n_0}{\lambda_1n_0(n_0+1)}
    \end{equation}

After deterministic oversampling,

    \begin{equation}\begin{aligned}
        \Delta_{{DI_2}_{\text{new}}}  &= \frac{n_1n_0 - \lambda_0(n_1 + n_2) - \dfrac{n_0}{10}}{\lambda_1n_0(10n_0+1)} \\ 
    & \approx_{\text{with } n_0, n_1 \text{big enough}} \frac{n_1n_0 - \lambda_0(n_1 + n_2) - n_0}{10\lambda_1n_0(n_0+1)} \\
    & = \frac{\Delta_{DI_2}}{10}
    \end{aligned}
    \end{equation}

\item Finally, changing an individual from $(S,\widehat{Y})=(1,1)$ to $(S,\widehat{Y})=(0,1)$ yields

\begin{equation}\begin{aligned}
        \Delta_{DI_2} = & \frac{(\lambda_0 + 1) (n_1 - 1)}{(\lambda_1 - 1)(n_0 + 1)} - \frac{\lambda_0   n_1}{\lambda_1 n_0} \\ \label{app:eq:change_4}
        & = \frac{(\lambda_0 + 1) (n_1 - 1)\lambda_1 n_0 - \lambda_0   n_1(\lambda_1 - 1)(n_0 + 1)}{\lambda_1 n_0(\lambda_1 - 1)(n_0 + 1)} \\
        & = \dfrac{\Big[-\lambda_0\lambda_1n_0 + n_1\lambda_1n_0 - \lambda_1n_0 \Big] - \Big[\lambda_0n_1\lambda_1 - \lambda_0n_1n_0 -\lambda_0n_1 \Big]}{\lambda_1 n_0(\lambda_1 - 1)(n_0 + 1)}
    \end{aligned}
    \end{equation}

Oversampling again reduces the variation proportionally,
\begin{equation}\begin{aligned}
        & \Delta_{{DI_2}_{\text{new}}}  = \dfrac{\Big[-10\lambda_0\lambda_1n_0 + 10n_1\lambda_1n_0 - \lambda_1n_0 \Big] - \Big[10\lambda_0n_1\lambda_1 - 10\lambda_0n_1n_0 -\lambda_0n_1 \Big]}{\lambda_1 n_0(10\lambda_1 - 1)(10n_0 + 1)} \\ 
    & \approx_{\text{with } \lambda_0, \lambda_1, n_0, n_1 \text{big enough}} \dfrac{\Big[-\lambda_0\lambda_1n_0 + n_1\lambda_1n_0 - \lambda_1n_0 \Big] - \Big[\lambda_0n_1\lambda_1 - \lambda_0n_1n_0 -\lambda_0n_1 \Big]}{10\lambda_1 n_0(\lambda_1 - 1)(n_0 + 1)} \\
    & = \frac{\Delta_{DI_2}}{10}
    \end{aligned}
    \end{equation}

\end{enumerate}

Across all possible modifications, we observe the same qualitative behavior: the variation $\Delta_{DI_2}$ decreases strictly as the dataset size increases. Achieving a proportional decrease therefore only requires sufficient oversampling.

In practice, oversampling was not required in any of our experiments, since the smallest dataset contained 500 individuals.

Consequently, we do not need to explicitly distinguish between the two definitions of Disparate Impact in the remainder of the paper. By applying oversampling if necessary, we can ensure that
\[
DI(f,\mathbb{P},S) = DI_2(f,\mathbb{P},S),
\]
since any case with $DI_2(f,\mathbb{P},S) < 1$ can be transformed into a configuration satisfying $DI_2(f,\mathbb{P},S) \leq 1$ after a single modification.

\subsection{Replacing key attributes and Wasserstein-minimizing sampling}
\label{app:sec:rep_key_attr} \label{app:sec:wass_sampling}

In this section, we precise how Disparate Impact (DI) can be increased using methods based on optimal transport. We can exchange between the 4 bins of points :$(Y=0, S=0), (Y=0, S=1), (Y=1, S=0)$ and $(Y=1, S=1)$, thus $4(4-1) = 12$ possible alterations. Due to the definition of DI, we can exclude the path from $(Y=1, S=0)$ to $(Y=0, S=0)$ and the path from $(Y=1, S=1)$ to $(Y=0, S=1)$ as it would decrease the DI, bringing the total to 10 possible transports. 

\begin{minipage}{\linewidth}%
\begin{algorithm}[H]
\begin{algorithmic}[2]
\State $\text{speed} \in \mathbb{N}^* ; 0 < \text{threshold} < 1$
\State $\text{b} = [|\{X|Y=1, S=1\}|, |\{X|Y=0, S=1\}|, |\{X|Y=1, S=0\}|, |\{X|Y=0, S=0\}|]$
\State $\text{DI} = \text{DI\_fct}(b)$
\State \text{swap\_possible} = $\{ Y_0S_0 ,Y_1S_1\}\rightarrow Y_0S_1, \{Y_0S_0\}\rightarrow Y_1S_1$ 
\State dic\_swap\_translation $= \{Y_0S_0 \rightarrow Y_1S_0 : [0, 0, 1, -1], Y_0S_0 \rightarrow Y_0S_1 : [0, 1, 0, -1], Y_1S_1 \rightarrow Y_1S_0 : [-1, 0, 1, 0] \}$ 
\State $\text{dic\_swap\_number} = {Y_0S_0 \rightarrow Y_1S_0 : 0, Y_0S_0 \rightarrow Y_0S_1 : 0, Y_1S_1 \rightarrow Y_1S_0 : 0 }$

\State $\text{DI}_n = [0,0,0] ; \text{Matrix}\_b = M_{(3,4)}(0)$
\While{DI < threshold}
  \State $i = 0$
    \For{$\text{swap} \in  \text{swap\_possible}$}:
    
        \State \text{b}\_n = \text{b} + dic\_swap\_translation[swap] \Comment{$Y_0S_0 \rightarrow Y_1S_0$ translation}
        \State $\text{Matrix}\_b[i,:] = \text{copy}(\text{b}\_n)$ \Comment{We keep in memory the bins}
        \State $\text{DI}_n[i] =  \text{DI\_fct}(\text{b}\_n)$
        \State $i = i + 1$
    \EndFor
    \State $j = \text{argmax}(\text{DI}_n)$
    \State dic\_swap\_done[swap\_possible[j]] = dic\_swap\_done[swap\_possible[j]] + speed \Comment{More information on speed discussed in next subsection}
    \State $b = b + \text{speed} * (\text{Matrix}\_b[j, :] - b)$
    \State DI $=$  DI\_fct(b) \Comment{Equal to $\text{DI}_n[j]$ only if speed $=1$}
\EndWhile
\State \Return dic\_swap\_number
\end{algorithmic}
\caption{\texttt{Replace($S$,$\widehat{Y}$)} non simplified algorithm}   \label{app:alg:DI_W_miti} 
\end{algorithm}
\end{minipage}

Moreover, If we consider the Wasserstein cost only on $(S,\widehat{Y})$, once again based on its definition, because it is more advantageous to have (Y=1,S=0) points instead of (Y=0, S=0), similarly for (Y=0, S=1) points instead of (Y=1, S=1), we only have 6 worthy alterations to consider instead of 10. Indeed, for instance, it would be suboptimal to transport a (Y=1, S=1) point to (Y=0, S=0) as moving it to (Y=1, S=0) would result in a higher DI with a lesser effort (with the cost is calculated only on $(S,\widehat{Y})$). 

Furthermore, transport between the bins $(Y=1,S=0)$ and $(Y=0,S=1)$ (i.e., between the two back points in Fig.~\ref{fig:Alteration}) can also be excluded from the optimal solution. Indeed, transport from $(Y=1,S=0)$ to $(Y=0,S=1)$ would both reduce the number of favorable outcomes (decreasing the numerator of the DI) and increase the number of unfavorable outcomes for the protected group (increasing the denominator), thus leading to a lower DI. In contrast, if the bin $(Y=0,S=1)$ move from $(Y=0,S=0)$, the DI is improved, as only the denominator increases while the numerator remains unchanged

To summarize, if the cost is calculated on $(S,\widehat{Y})$, then we theoretically only have 4 moves to consider, the arrows in Fig. \ref{fig:Alteration}. The arrow from (Y=0, S=0) toward (Y=0, S=1) is doted for the following reason: in practice, for the simulated dataset presented in Section \ref{app:sec:simu}, this transport was never optimal meaning not the one which increases the DI the most compared to other transports. The most rewarding one was usually from the transport from (Y=0,S=0) to (Y=1,S=0). This leads us to write the Alg.~\ref{app:alg:DI_W_miti} which is the less concise version of Alg.~\ref{alg:Replace}. A notable difference between the two is that Alg.~\ref{app:alg:DI_W_miti} has a \texttt{speed} parameter which express a trade-off performance rapidity as explained in Section~\ref{app:sec:cost}.

\paragraph{Termination analysis}
At every iteration of the while loop, the DI is strictly increasing. moreover, the number of iterations is limited by the number of points $|\{X|Y=1, S=1\}|$ and $|\{X|Y=0, S=0\}|$. 
In the extreme case where no transport would be possible (either of these sets is empty if $|\{X|Y=1, S=1\}|=0$ or $|\{X|Y=0, S=0\}|=0$) the algorithm could attempt to increase DI indefinitely (towards $+\inf$). This is crucial here to consider the Disparate impact definition of initial minority ($S=0$) over majority ($S=1$) and not the min over max definition: in the min/max definition, the DI is always within 0 and 1 ; in contrary, the initial minority over majority starts at less than 1, but can increase over 1 if the direction of the potential discrimination changed ($S=1$ becomes the class which might be discriminated against by the model). This ensures that the algorithm necessarily terminates.

\paragraph{Objective analysis} 
The condition of the while loop is precisely aligned with the objective of our problem. Consequently, exiting the loop implies that a solution has been found. Finally, a more challenging question concerns the optimality of the solution returned by the algorithm. We leave this question open and do not provide a formal guarantee of optimality.

\subsection{Wasserstein gradient guided method}
\label{app:sec:wass_grad}

To complete the method's explanation, we need to clarify how to choose $t_0$ and $t_1$ and $\lambda$:
\begin{enumerate}
    \item The choice of $t_0$ and $t_1$ is similar as in section \ref{sec:KL}: either the balanced case where $\delta_0 = \delta_1$ or the proportional case where $\frac{\delta_0}{n_0} = \frac{\delta_1}{n_1}$.
    \item $\lambda$ is a constraint regulation coefficient, meaning that the bigger $\lambda$ is, the more the optimization solution will take into account the constraint \\ $\int_{x \in E | S = s} f(x)dQ(x) \leq t_{s}$. And consequently, the bigger $\lambda$ is, the farther the solution will be from the original distribution. 
    Therefore, we start by solving \eqref{eq:prblm_wass_impl} with a low $\lambda$, and we increase it until the constraint is respected.
\end{enumerate}

The constraints are imposed separately on the squared Wasserstein distances $W_2^2(Q_{n,1}, Q_{t,1})$ and $W_2^2(Q_{n,0}, Q_{t,0})$. The inequality $W_2^2(\pi Q_{n,1} + (1-\pi)Q_{n,0} , \pi Q_{t,1} + (1-\pi)Q_{t,0}) \leq \pi W_2^2(Q_{n,1}, Q_{t,1}) + (1 - \pi) W_2^2(Q_{n,0}, Q_{t,0})$ provides an upper bound on the overall distance between the two samples. The proof of this result is deferred to \ref{app:sec:proof_ineq}.

\begin{algorithm}[H]
\caption{Fairwashing using Monge Kantorovich constrained projection algorithm \texttt{Grad}}
\label{app:alg:MK_c_proj}
\begin{algorithmic}[1]
\Require Neural network $f$, data $Z_0$, sensitive attribute $S \in \{0,1\}^n$, prediction threshold $\tau$, desired DI threshold $t$, learning rate $\eta$, constraint weight $\lambda$, delta type ($\in \{\text{\texttt{balanced}, \texttt{proportional}}\}$
\Ensure Updated samples $Z$ minimizing $\|Z - Z_0\|^2$ while satisfying $\mathrm{DI}(f(Z), S) \geq t$
\vspace{1mm}
\State \textbf{Compute:} $\widehat{Y} \gets \mathbb{I}[f(Z_0) > \tau]$ \Comment{where $\mathbb{I}$ is the indicator function}
\State Compute $P_0 = \mathbb{E}[\widehat{Y} \mid S=0]$, $P_1 = \mathbb{E}[\widehat{Y} \mid S=1]$, $n_1 = \mathbb{E}[ S=1]$, $n_0 = \mathbb{E}[ S=0]$
\State Compute $\delta_s$ according to delta type \Comment{Done following Prop.\ref{th:FWKL}}

\State Set new target rates:
\[
\tilde{P}_1 = P_1 - \delta_1/n_1, \quad
\tilde{P}_0 = P_0 + \delta_0/n_0
\]

\For{$s \in \{0,1\}$}
    \State Initialize $\lambda^{(s)} \gets \lambda$
    \While{ ($\mathbb{E}[\widehat{Y}^{(s)}] < \tilde{P}_s$) $\lor$ ($\mathbb{E}[\widehat{Y}^{(s)}]> \tilde{P}_s$ $\land$ $s=1$)}
        \State Initialize $Z^{(s)}_i \gets Z_0^{(s)}$, $\eta_i \gets \eta$ \Comment{where $Z_0^{(s)}$ is the subset of inputs with $S = s$}
        \For{$i \in 1,\cdots,10$}
            \State Iteration of gradient step:
            \[
            \nabla = 2(Z^{(s)}_i - Z_0^{(s)}) + \lambda^{(s)} \cdot \nabla_{Z} f(Z^{(s)}_i) \cdot d_s
            \]
            where $d_s = \begin{cases} +1 & \text{if } s=0 \\ -1 & \text{if } s=1 \end{cases}$\Comment{Gradient choice following Thm.~\ref{thm:empirical}}
            \[
            Z^{(s)}_i \gets Z^{(s)}_i - \eta_i \cdot \nabla
            \]
            \State Recompute predictions $\widehat{Y}^{(s)}_i = \mathbb{I}[f(Z^{(s)}_i) > \tau]$
            \State $\eta_i \gets \eta_i / 1.2$ \Comment{Planning strategies could improve the performance, 1.2 was what we founded worked the best in practice (following the choice of the coefficient multiplying $\lambda^{(s)}$)}
        \If{\texttt{1D-transport variant}}
            \State Project each feature of $Z^{(s)}_i$ to its closest achievable value
        \EndIf
        \If{$\mathbb{E}[\widehat{Y}^{(s)}_i] < \tilde{P}_s$ (or $> \tilde{P}_s$ for $s=1$)} 
            \State \textbf{Break} Exit for and while loop
        \EndIf
        \EndFor
        \State Update : $\lambda^{(s)} \gets 1.2 \times\lambda^{(s)}$ \Comment{The solution of the optimization problem with this $\lambda$ is not within the constrained space (or we did not converge towards it fast enought at least); hence we increase the $\lambda$ progressively. Note that the $1.2$ was what we founded worked best in practice (trade-off between precision with lower value and fast computation), however further tuning would be relevant.}
        
    \EndWhile
    \State Compute perturbation $T^{(s)} = Z^{(s)} - Z_s$
\EndFor
\State Assemble final perturbation $T$ such that:
\[
T_i = 
\begin{cases}
T^{(0)}_i & \text{if } S_i = 0 \text{ and } \widehat{Y}_i = 0 \\
T^{(1)}_i & \text{if } S_i = 1 \text{ and } \widehat{Y}_i = 1 \\
0 & \text{otherwise}
\end{cases}
\]
\State \Return $Z = Z_0 + T$
\end{algorithmic}
\end{algorithm}

Alg.~\ref{app:alg:MK_c_proj}, which is a simplified version of the true algorithm (code available on \href{https://github.com/ValentinLafargue/Inspection}{Github}), explains the main ideas being: 
\begin{enumerate}
    \item We find the target probabilities for each subgroup of $s$
    \item We treat each $Q_{n,s} := Q_n(\cdot \mid S = s)$ separately
    \item The gradient steps stem from Theorem~\ref{thm:empirical}
    \item We start with a small constraint weight and increase it progressively
\end{enumerate}

The elements present in our code but which we did not include in Alg.~\ref{app:alg:MK_c_proj} for visibility are mostly computational optimizations. For instance, we did not compute the gradient on neither the points whose network decision we would not modify $Z_i$ (i.e., with $\text{if } S_i = 0 \text{ and } \widehat{Y}_i = 1$) nor on points $Z_i^{(s)}$ whose $\widehat{Y}_i^{(s)}$ are already modified. We also kept streakily only the minimum number of modification necessary: some gradient step would change the network decision of multiple points at the same time and without this process our result would not be tight regarding $\tilde{P}_1, \tilde{P}_0 $ (as we would have changed individuals' outcome more than necessary) and thus overachieving $\mathrm{DI}(f(Z), S) \geq t$ which is not beneficial in our use case where we highlighted the trade-off between fairness correction and distribution shift.

\subsection{Costs of the methods, solutions and tests}
\label{app:sec:cost}

\begin{table}[tb!]
        \caption{Time cost analysis of the methods, note that every estimation depends on the original dataset, its Disparate Impact and the DI constraint. Time estimation given for a dataset size of 20k individuals.}
      \label{app:tab:time}
  \centering
  \resizebox{\textwidth}{!}{\begin{tabular}{l l l}
    \toprule
     Methods & Summary & Solution\\
    \midrule
    $M_{W(X,S,\widehat{Y})}$            & 3–10 minutes   & Trade-off possible \\
    \texttt{Replace($S$,$\widehat{Y}$)}           & $\leq 2$ minutes & Trade-off possible \\
    \texttt{Entropic\_b}   / \texttt{Entropic\_p}     & $\leq 1$ minutes  & \\
    \texttt{Grad\_p}       / \texttt{Grad\_b}         & 3–15 minutes, depends on $\lambda$ and NN architecture & Trade-off possible \\
    \texttt{Grad\_p(1D-t)} / \texttt{Grad\_b(1D-t)}   & 3–20 minutes, depends on $\lambda$ and NN architecture & Trade-off possible \\
    \bottomrule
    \vspace{-0.7cm}
  \end{tabular}}

\end{table}

\begin{table}[tb!]
  
  \centering
  \caption{Time analysis done during our Highest undetected achievable DI per datasets and methods}.
  \label{app:tab:test_time}
\begin{tabular}{llc}
\toprule
Sample size & Test performed & Average testing time (second)\\
\midrule
500 & $\text{DI}(\widehat{\mathcal{Q}}_m) \geq \text{DI}(Q_t) $ & 0.00 \\
& $\KL(S,\widehat{Y})$   & 0.00 \\
& $\KL(X,S,\widehat{Y})$ & 0.78 \\
& $W(S,\widehat{Y})$    & 0.29 \\ 
& $W(X,S,\widehat{Y})$  & 0.48 \\
\midrule
1000 & $\text{DI}(\widehat{\mathcal{Q}}_m) \geq \text{DI}(Q_t) $ & 0.00 \\
& $\KL(S,\widehat{Y})$   & 0.00 \\
& $\KL(X,S,\widehat{Y})$ & 0.85 \\
& $W(S,\widehat{Y}) + W(X,S,\widehat{Y})$ & 1.57 \\
\midrule
2000 & $\text{DI}(\widehat{\mathcal{Q}}_m) \geq \text{DI}(Q_t) $ & 0.00 \\
& $\KL(S,\widehat{Y})$   & 0.02 \\
& $\KL(X,S,\widehat{Y})$ & 3.13 \\
& $W(S,\widehat{Y})$    & 4.93 \\
& $W(X,S,\widehat{Y})$  & 15.51 \\
\midrule
4000 & $\text{DI}(\widehat{\mathcal{Q}}_m) \geq \text{DI}(Q_t) $ & 0.00 \\
& $\KL(S,\widehat{Y})$   & 0.02 \\
& $\KL(X,S,\widehat{Y})$ & 3.34 \\
& $W(S,\widehat{Y})$    & 9.58 \\
& $W(X,S,\widehat{Y})$  & 32.30 \\
\bottomrule
\vspace{-0.7cm}
\end{tabular}

\end{table}

\subsubsection{Time.} In Table \ref{app:tab:time}, we wrote Trade-off possible for the methods which might take a more than a day to run with millions of individuals. The methods $M_{W(X,S,\widehat{Y})}$ and \texttt{Replace($S$,$\widehat{Y}$)} evaluate at each step amongst 3 or 4 possibilities which is the optimal to take, we can only evaluate once for more step at the same time for both methods, this becomes a trade-off between speed and precision, this is what we mean by trade-off possible for those methods. Moreover, we can also think about a trade-off about the number of transport mapping to consider, as explained in the Section.~\ref{app:sec:wass_grad}. 

For the \texttt{Grad} variant methods, we do not anticipate any changes to the model architecture. However, if inference from the neural network is computationally expensive, the overall cost of the method will also be high. Developing an efficient solution to this issue remains an open challenge. However, with tabular data model's number of parameters tends to be controllable, and thus in our experiments the reason of such a long time compute time (relative to the number of individual) was because we optimized for the $\lambda$ parameter. We remind that to have the best results we start with a very small $\lambda$ which we progressively increase ; we thus can simply initialize the algorithm with a bigger $\lambda$ to save computing time, another speed precision trade-off.

The results in Table~\ref{app:tab:test_time} were obtained through the following procedure. For each sample, we recorded: (1) the total execution time of the testing pipeline, and (2) the reason the pipeline stopped. Since each sample must pass all five tests, the pipeline halts as soon as one test is failed. Based on our prior expectations regarding the relative runtime of the tests. To isolate the runtime of each individual test, we subtracted the mean runtime of the preceding tests from the total time observed at the stopping point.
The results show that while all tests are fast for small sample sizes (e.g., 500 samples), the tests based on Wasserstein distances (in particular $W(X,S,\widehat{Y})$) are the most time-consuming.

\begin{table}[ht!]
  \caption{Memory cost analysis of the methods for a $N \times J$ dataset.}
  \label{app:tab:memory}
  \centering
  \resizebox{\textwidth}{!}{\begin{tabular}{l l l}
    \toprule
     Methods & Summary & Solution\\
    \midrule
    $M_{W(X,S,\widehat{Y})}$            & $N \times N$ distance matrix & \\
    \texttt{Replace($S$,$\widehat{Y}$)}         & Negligible & Trade-off possible \\
    \texttt{Entropic\_b}   / \texttt{Entropic\_p}     & Negligible  & \\
    \texttt{Grad(b/p)/(1D)}                  & NN gradient to compute on at worse on $N$ ind & Batch approach \\
    \bottomrule
  \end{tabular}}

\end{table}

\subsubsection{Memory} We consider only the \texttt{Grad} variant methods to potentially pose memory-related issues. Although it would be natural to adapt these methods to operate in a batch-wise manner, we did not implement such an approach in our current work.

\section{Optimization result values}  

\label{app:sec:result}
\subsection{Wasserstein distance}

\begin{table}[h!]
  \caption{Wasserstein distance manipulation cost of the fairwashing methods ($\text{DI}(Q_t) \geq 0.8$), cost calculated on the projected dataset : $W(Q_{n}, Q_{t})$ with the original dataset $Q_n$ and $Q_t = f(Q_n)$ with $f$ the fairwashing method}
  \label{app:fum_wass_result_A}
  \centering
  \resizebox{\textwidth}{!}{\begin{tabular}{lll lll lll}
    \toprule
    & \multicolumn{8}{c}{Unbiasing Methods}                   \\
     \cmidrule(r){2-9}
    Dataset & Grad\_p & Grad\_b & Grad\_p\_1D & Grad\_b\_1D & Rep $(S,\widehat{Y})$ & $M_{W(X,S,\widehat{Y})}$ & Entr\_b & Entr\_p \\
    \midrule
    ADULT & 0.10 & 0.08 & 0.13 & 0.09 & \textbf{0.05} & \textbf{0.06} & 0.28 & 0.35 \\
    EMP   & 0.18 & 0.10 & 0.18 & 0.10 & \textbf{0.06} & 0.08 & 0.22 & 0.37 \\
    INC   & 0.01 & 0.01 & 0.01 & 0.01 & 0.01 & 0.01 & 0.01 & 0.01 \\
    MOB   & 0.21 & 0.06 & 0.23 & 0.08 & \textbf{0.03} & 0.05 & 0.18 & 0.64 \\
    PUC   & 0.21 & \textbf{0.13} & 0.22 & 0.14 & \textbf{0.12} & 0.16 & 0.33 & 0.48 \\
    TRA   & 0.02 & 0.02 & 0.02 & 0.02 & \textbf{0.01} & \textbf{0.01} & 0.03 & 0.03 \\
    BAF   & 0.01 & \textbf{0.00} & 0.02 & 0.01 & \textbf{0.00} & 0.01 & 0.02 & 0.05 \\
    \bottomrule
    \vspace{-0.7cm}
  \end{tabular}}

\end{table}

\clearpage

\begin{table}[h!]
  \caption{Wasserstein distance manipulation cost of the fairwashing methods($\text{DI}(Q_t) \geq 0.8$), cost calculated on the projected dataset : $W(Q_{n, (S,\widehat{Y})}, Q_{t, (S,\widehat{Y})})$ (where $X$ is excluded), with the original dataset $Q_n$ and $Q_t = f(Q_n)$ with $f$ the fairwashing method}
  \label{app:fum_wass_result_SY}
  \centering
  \resizebox{\textwidth}{!}{\begin{tabular}{lll lll lll}
    \toprule
    & \multicolumn{8}{c}{Unbiasing Methods}                   \\
     \cmidrule(r){2-9}
    Dataset & Grad\_p & Grad\_b & Grad\_p\_1D & Grad\_b\_1D & Rep $(S,\widehat{Y})$ & $M_{W(X,S,\widehat{Y})}$ & Entr\_b & Entr\_p \\
    \midrule
    ADULT & 0.09 & 0.08 & 0.09 & 0.08 & \textbf{0.05} & \textbf{0.05} & 0.08 & 0.09 \\
    EMP   & 0.18 & 0.10 & 0.18 & 0.10 & \textbf{0.06} & \textbf{0.06} & 0.10 & 0.18 \\
    INC   & 0.01 & 0.01 & 0.01 & 0.01 & 0.01 & 0.01 & 0.01 & 0.01 \\
    MOB   & 0.18 & 0.06 & 0.18 & 0.06 & \textbf{0.03} & \textbf{0.03} & 0.06 & 0.18 \\
    PUC   & 0.21 & 0.13 & 0.21 & 0.13 & 0.12 & \textbf{0.10} & 0.13 & 0.21 \\
    TRA   & 0.02 & 0.02 & 0.02 & 0.02 & \textbf{0.01} & \textbf{0.01} & 0.02 & 0.02 \\
    BAF   & 0.01 & 0.00 & 0.01 & 0.00 & 0.00 & 0.00 & 0.00 & 0.00 \\
    \bottomrule
    \vspace{-0.7cm}
  \end{tabular}}

\end{table}

\subsection{Kullback-Leibler divergence}

\begin{table}[h!]
  \caption{KL divergence manipulation cost of the fairwashing methods ($\text{DI}(Q_t) \geq 0.8$), cost calculated on the projected dataset : $\text{KL}(Q_{n}, Q_{t})$ with the original dataset $Q_n$ and $Q_t = f(Q_n)$ with $f$ the fairwashing method}
  \label{app:fum_kl_result_A}
  \centering
  \resizebox{\textwidth}{!}{\begin{tabular}{lll lll lll}
    \toprule
    & \multicolumn{8}{c}{Unbiasing Methods}                   \\
     \cmidrule(r){2-9}
    Dataset & Grad\_p & Grad\_b & Grad\_p\_1D & Grad\_b\_1D & Rep $(S,\widehat{Y})$ & $M_{W(X,S,\widehat{Y})}$ & Entr\_b & Entr\_p \\
    \midrule
    ADULT & $\infty$ & $\infty$ & $\infty$ & $\infty$ & $\infty$ & 0.03 & \textbf{0.02} & 0.03 \\
    EMP   & $\infty$ & $\infty$ & $\infty$ & $\infty$ & $\infty$ & \textbf{0.04} & \textbf{0.04} & 0.07 \\
    INC   & $\infty$ & $\infty$ & $\infty$ & $\infty$ & $\infty$ & 0.00 & 0.00 & 0.00 \\
    MOB   & $\infty$ & $\infty$ & $\infty$ & $\infty$ & $\infty$ & \textbf{0.02} & 0.03 & 0.17 \\
    PUC   & $\infty$ & $\infty$ & $\infty$ & $\infty$ & $\infty$ & 0.06 & 0.07 & 0.10 \\
    TRA   & $\infty$ & $\infty$ & $\infty$ & $\infty$ & $\infty$ & 0.00 & 0.00 & 0.00 \\
    BAF   & $\infty$ & $\infty$ & $\infty$ & $\infty$ & $\infty$ & 0.00 & 0.00 & 0.00 \\
    \bottomrule
    \vspace{-0.7cm}
  \end{tabular}}

\end{table}

\begin{table}[h!]
  \caption{KL divergence manipulation cost of the fairwashing methods ($\text{DI}(Q_t) \geq 0.8$), cost calculated on the projected dataset : $\text{KL}(Q_{n, (S,\widehat{Y})}, Q_{t, (S,\widehat{Y})})$ (where $X$ is excluded) with the original dataset $Q_n$ and $Q_t = f(Q_n)$ with $f$ the fairwashing method}
  \label{app:fum_kl_result_SY}
  \centering
  \resizebox{\textwidth}{!}{\begin{tabular}{lll lll lll}
    \toprule
    & \multicolumn{8}{c}{Unbiasing Methods}                   \\
     \cmidrule(r){2-9}
    Dataset & Grad\_p & Grad\_b & Grad\_p\_1D & Grad\_b\_1D & Rep $(S,\widehat{Y})$ & $M_{W(X,S,\widehat{Y})}$ & Entr\_b & Entr\_p \\
    \midrule
    ADULT & 0.02 & 0.02 & 0.02 & 0.02 & 0.03 & 0.03 & 0.02 & 0.03 \\
    EMP   & 0.06 & \textbf{0.03} & 0.06 & \textbf{0.03} & 0.04 & 0.04 & 0.04 & 0.07 \\
    INC   & 0.00 & 0.00 & 0.00 & 0.00 & 0.00 & 0.00 & 0.00 & 0.00 \\
    MOB   & 0.12 & \textbf{0.03} & 0.12 & 0.03 & \textbf{0.02} & \textbf{0.02} & \textbf{0.03} & 0.17 \\
    PUC   & 0.09 & \textbf{0.06} & 0.09 & \textbf{0.06} & 0.08 & \textbf{0.07} & \textbf{0.07} & 0.10 \\
    TRA   & 0.00 & 0.00 & 0.00 & 0.00 & 0.00 & 0.00 & 0.00 & 0.00 \\
    BAF   & 0.00 & 0.00 & 0.00 & 0.00 & 0.00 & 0.00 & 0.00 & 0.00 \\
    \bottomrule
  \end{tabular}}

\end{table}

\section{Ablation study on the MMD test}

We report the results in Table~\ref{app:tab:highest_undetected_unbiasing_woMMD}, which presents the highest DI achieved by samples that remained undetected from the statistical tests based on KL divergence, Wasserstein distance, and the Kolmogorov–Smirnov test. This table corresponds to Table~\ref{tab:highest_undetected_unbiasing}, but excludes the two tests based on the MMD distance. 

From Table~\ref{app:tab:highest_undetected_unbiasing_woMMD}, we observe that excluding the MMD tests had negligible impact on detection outcomes. The only notable difference arises in the BAF dataset with a 20\% sampling rate, where the achieved DI is slightly higher. We also point out that, due to the inherent randomness in sampling (100 random samples are drawn for each combination of dataset, fairwashing method, sample size, and target $\text{DI}(Q_t)$), we occasionally found samples that passed all seven tests and exhibited marginally higher DI than those evaluated with only five tests. These cases are indicated by a ‘+’ symbol in parentheses in Table~\ref{app:tab:highest_undetected_unbiasing_woMMD}.

\begin{table*}[h!]
    \centering
    \caption{Highest undetected (without the MMD-based statistical tests) achievable Disparate Impact for each dataset, sample size (S Size) and fairwashing method. The symbol -- indicates that some methods couldn't reach a DI improvement. To emphasize the best method to use in order to deceive the auditor, we put the DI achieved in bold when one or two over-performed the others. The number in parentheses are here to indicate the difference between those results and the results obtained with the MMD-based tests (Result without MMD - Result with).}
    \label{app:tab:highest_undetected_unbiasing_woMMD}
    \begin{tabular}{ll c cccc}
\toprule
Dataset & Original & S size (\%) & Rep $(S, \widehat{Y})$ & Entr\_b & Entr\_p & $M_{W(X,S,\widehat{Y})}$ \\
\midrule
ADULT & 0.30 & 10 & 0.45(-0.05) & 0.53 (-0.01) & 0.55 (+0.03) & \textbf{0.54} (+0.01) \\
      &      & 20 & 0.38(-0.03) & \textbf{0.43}(+0.01) & 0.42(+0.01) & \textbf{0.43}(+0.01) \\
EMP   & 0.30 & 10 & --   & 0.38(+0.03) & \textbf{0.39}(+0.03) & \textbf{0.39}(+0.02) \\
      &      & 20 & --   & \textbf{0.36}(+0.02) & 0.35(-0.01) & \textbf{0.36}(+0.01) \\
INC   & 0.67 & 10 & 0.88 & 0.95(+0.01) & 0.95(+0.01) & 0.95(+0.02) \\
      &      & 20 & 0.83 & 0.84(+0.01) & 0.84 & 0.84 \\
MOB   & 0.45 & 10 & 0.54(+0.01) & 0.53(+0.01) & 0.51 & \textbf{0.55}(+0.02) \\
      &      & 20 & 0.48(-0.01) & \textbf{0.50} & 0.49 & \textbf{0.50} \\
PUC   & 0.32 & 10 & --   & \textbf{0.36}(+0.03) & \textbf{0.36}(+0.01) & 0.35 \\
      &      & 20 & --   & --   & --   & --   \\
TRA   & 0.69 & 10 & 0.76(-0.03) & 0.84(+0.01) & 0.84 & 0.84 \\
      &      & 20 & 0.71(-0.06) & 0.80(+0.01) & 0.80(+0.01) & \textbf{0.81} \\
BAF   & 0.35 & 10 & --   & 1    & 1    & 1    \\
      &      & 20 & --   & 0.83(+0.06) & 0.84(+0.05) & \textbf{0.85}(+0.06) \\
\bottomrule
\end{tabular}    
\end{table*}

\section{Fraud detection}
\subsection
  [Distribution of tries before acceptance of H0]
  {Distribution of tries before acceptance of $\mathcal{H}_0$}
\label{app:sec:tries}

\begin{figure}[tb!]
    \centering
    \begin{subfigure}[t]{0.45\textwidth}
        \centering
        \includegraphics[height=2in]{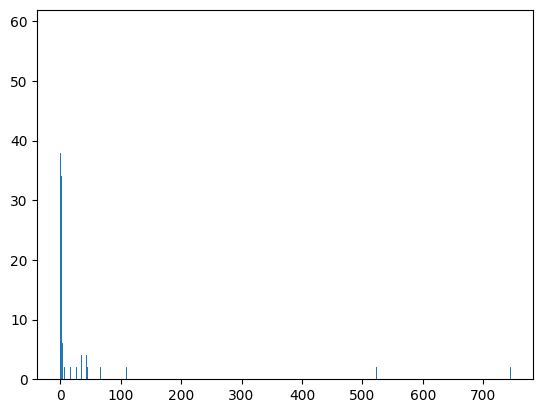}
        \caption{Complete distribution}
        \label{fig:Dist_HO_1}
    \end{subfigure}
    \hfill
    \begin{subfigure}[t]{0.45\textwidth}
        \centering
        \includegraphics[height=2in]{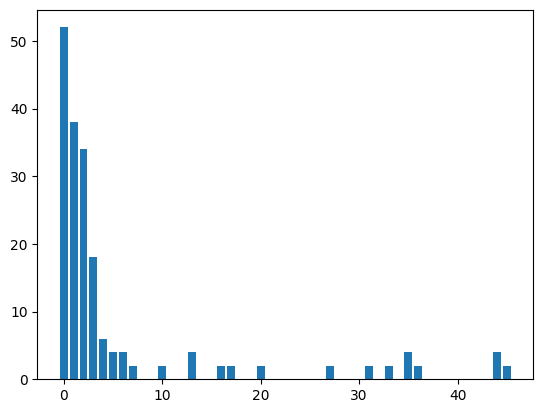}
        \caption{Zoom on under 50}
        \label{fig:Dist_HO_2}
    \end{subfigure}
    \caption{Distribution of number of tries to find an accepted sample for $\mathcal{H}_0$ for the statistical test KS or $\text{KL}(S,\widehat{Y})$ with a maximum of 1000 tries per configuration (method, dataset, test) for all datasets.}
    \label{fig:Distr_H0}
\end{figure}

As shown in Fig~\ref{fig:Distr_H0}, taking only 30 or 50 samples instead of 1000 gives us respectively 73\% or 78\% accuracy for the tests. This is arguably not that high, however knowing that we would have needed the combination of 5 statistical tests not to reject the sample in our use case, it still gives us a good approximation to whether the test have a chance not to be rejected (as it is harder not to be rejected by none of the 5 tests than only the individuals one).

\subsection{Highest undetected achievable Disparate Impact probabilities and additional graph} \label{app:sec:Highest}

\begin{figure}[h!]
    \centering
    \begin{minipage}{0.47\textwidth}
        \centering
\includegraphics[height=1.8in]{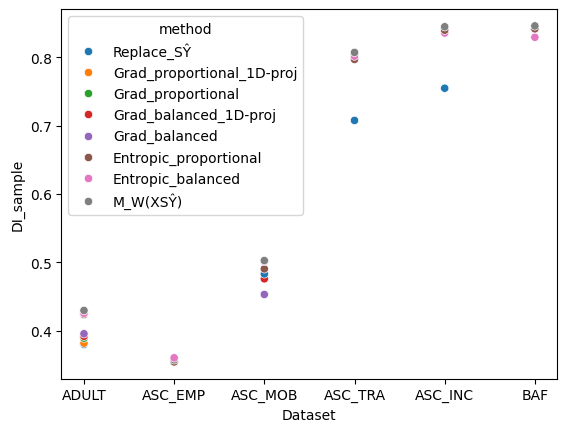}
\caption{Highest achieved DI for all Datasets and methods (when they improve the original DI), with sample size of 20\% of the dataset.}
\label{app:fig:highest_20}
    \end{minipage}
    \hfill
    \begin{minipage}{0.47\textwidth}
        \centering
\includegraphics[height=1.8in]{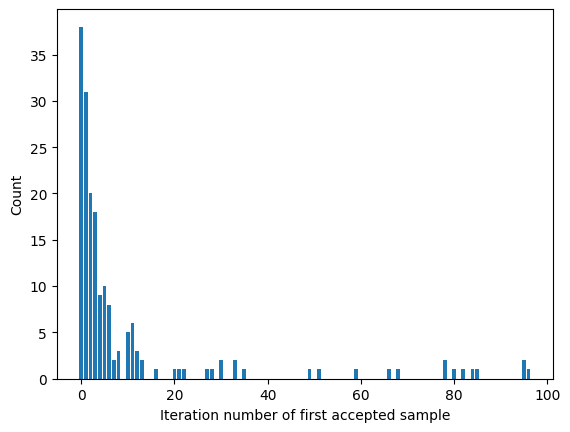}
\caption{Distribution of the number of sample tried before first accepted one for all datasets.}
    \label{app:fig:distr_highest}
    \end{minipage}
\end{figure}

We add details to Table~\ref{tab:highest_undetected_unbiasing} results, particularly its stability towards the number of sampling tries. First, if we tried 10, 20 and 30 samples compared to 100 we would have had respectively 95\%, 97\% and 98\% of result correspondence. Secondly, in configurations where a fairer falsely compliant sample was found, it was generally around the $11^{th}$ sample, while the median was equal to 4. (See Fig.~\ref{app:fig:distr_highest})

\section{Disclosure of LLM Use}
Large Language Models (LLMs) were used in a limited capacity during the preparation of this paper. Their use was restricted to (i) spelling and phrasing assistance (to support a dyslexic co-author), and (ii) suggesting improvements to Python scripts for graph generation and visualization. No part of the scientific content, analyses, or conclusions was produced by LLMs.

\end{document}